\pdfoutput=1
\documentclass[conference]{IEEEtran}
\usepackage{cite}
\usepackage{textcomp}
\usepackage[pdftex,dvipsnames]{xcolor}
\usepackage{booktabs}       
\usepackage{amsfonts}       
\usepackage{nicefrac}       
\usepackage{microtype}      
\usepackage[para]{footmisc}
\usepackage{pstricks}
\usepackage{amsmath,amssymb,bm}

\def\eqref#1{equation~\ref{#1}}

\def\1{\bm{1}}

\newcommand{\test}{\mathcal{D_{\mathrm{test}}}}

\def\mC{{\bm{C}}}

\def\mG{{\bm{G}}}

\def\mL{{\bm{L}}}

\def\mN{{\bm{N}}}

\def\mP{{\bm{P}}}

\def\mS{{\bm{S}}}

\def\mU{{\bm{U}}}

\def\mW{{\bm{W}}}

\DeclareMathAlphabet{\mathsfit}{\encodingdefault}{\sfdefault}{m}{sl}
\SetMathAlphabet{\mathsfit}{bold}{\encodingdefault}{\sfdefault}{bx}{n}

\newcommand{\E}{\mathbb{E}}

\newcommand{\R}{\mathbb{R}}

\usepackage{amsthm,dsfont}
\usepackage{graphicx,here,comment}
\usepackage{caption}

\usepackage{algorithm,algorithmicx}
\usepackage[noend]{algpseudocode}

\usepackage{multirow}
\usepackage{bigints}
\usepackage{xargs}                      
\usepackage{enumerate}

\usepackage{ mathrsfs }
\usepackage{subcaption}
\usepackage{wrapfig}

\usepackage{hyperref}
\newtheorem{theorem}{Theorem}
\newtheorem{corollary}{Corollary}

\theoremstyle{definition}
\newtheorem{definition}{Definition}

\newcommand*\Let[2]{\State #1 $\gets$ #2}

\newcommand\restr[2]{{
		\left.\kern-\nulldelimiterspace 
		#1 
		\vphantom{\big|} 
		\right|_{#2} 
}}

\newcommand{\rarrow}[1]{\buildrel #1 \over \longrightarrow}

\def\BibTeX{{\rm B\kern-.05em{\sc i\kern-.025em b}\kern-.08em
    T\kern-.1667em\lower.7ex\hbox{E}\kern-.125emX}}
\begin{document}

\title{Learning Representations using Spectral-Biased Random Walks on Graphs
}

\author{\IEEEauthorblockN{Charu Sharma, Jatin Chauhan, Manohar Kaul}
\IEEEauthorblockA{\textit{Dept. of Computer Science \& Engineering} \\
\textit{Indian Institute of Technology Hyderabad}\\
Kandi, Sangareddy, India \\
Email: \{cs16resch11007, cs17btech11019, mkaul\}@iith.ac.in}

}

\maketitle

\begin{abstract}
Several state-of-the-art neural graph embedding methods are based on short random walks (stochastic processes) because of their ease of computation, simplicity in capturing complex local graph properties, scalability, and interpretibility. In this work, we are interested in studying how much a probabilistic bias in this stochastic process affects the quality of the nodes picked by the process. In particular, our biased walk, with a certain probability, favors movement
  towards nodes whose neighborhoods bear a structural resemblance to the current node's neighborhood. We succinctly capture this neighborhood as a probability measure based on the spectrum of the node's neighborhood subgraph represented as a normalized Laplacian matrix.
	We propose the use of a paragraph 
	vector model with a novel Wasserstein regularization term.
	We empirically evaluate our approach against several state-of-the-art node embedding techniques on a wide variety of real-world datasets and demonstrate that our proposed method significantly improves upon existing methods on both link prediction and node classification tasks. 	

\end{abstract}


\section{Introduction}
Graph embedding methods have gained prominence in a wide variety of tasks including pattern recognition~\cite{monti2017}, low-dimensional embedding~\cite{belkin2001,brand2003}, node classification~\cite{grover2016node2vec,abu2018watch,perozzi2014deepwalk}, and link prediction~\cite{zhang2018link,abu2018watch}, to name a few. In machine learning, the task of producing graph embeddings entails
capturing \emph{local} and \emph{global} graph statistics and encoding them as vectors that best preserve these statistics in a computationally efficient manner.
Among the numerous graph embedding methods, we focus on \emph{unsupervised} graph embedding models, which can be broadly classified as \emph{heuristics} and \emph{random walk} based models.

\emph{Heuristic based models} compute node similarity scores based on vertex neighborhoods and are further categorized based on the 
maximum number of $k$-hop neighbors they consider around each vertex\footnote{``vertex" and ``node" will be used interchangeably.}. 
Recently, Zhang et. al.~\cite{zhang2018link} proposed a graph neural network (GNN) based framework that required enclosing subgraphs around each edge in the graph. They showed that most higher order heuristics ($k>2$) are a special case of their proposed $\gamma$-\emph{decaying heuristic}. While their method outperforms the heuristic based methods on link prediction, it nevertheless computes all walks of length at most $k$ (i.e., the size of the neighborhood) around each edge, which is quite prohibitive and results in being able to only process small and sparse graphs.

In comparison, \emph{random walk based models} are scalable and have been shown to produce good quality embeddings. These methods generate several short random walks originating from each node and then embed a pair of nodes close to one another in feature space, if they co-occur more frequently in several such walks. This is achieved by treating each random walk as a \emph{sequence of words} appearing in a sentence and feeding this to a word-embedding model like word2vec~\cite{mikolov2013}. Deepwalk~\cite{perozzi2014deepwalk} first proposed this approach, after which many works~\cite{grover2016node2vec,Ribeiro2017,Chen2018} followed suit. 
Recently, WYS~\cite{abu2018watch} presented a \emph{graph attention} (GAT) model that is based on simple random walks and learning a \emph{context distribution}, which is the probability of encountering a vertex in a variable sized context window, centered around a fixed anchor node.
An important appeal of random walks is that they concisely capture the underlying graph structure surrounding a vertex. Yet, further important structure remains uncaptured. For example, heuristic methods rely on the intuition that vertices with similar $k$-hop neighborhoods should also be closer in feature space, while simple random walks cannot guarantee the preservation of any such \emph{grouping}. In WYS, under certain settings of the context window size, vertices with structurally similar neighborhoods can easily be omitted and hence overlooked.

In our work, we incorporate such a grouping of structurally similar nodes \emph{directly} into our random walks. Our novel methodology opens avenues to a richer class of \emph{vertex grouping} schemes. To do so, we introduce \emph{biased random walks}~\cite{Morena2014,Azar1996} that \emph{favor}, with a certain probability, moves to adjacent vertices with similar $k$-hop neighborhoods. 

First, we capture the structural information in a vertex's neighborhood by assigning it a \emph{probability measure}. This is achieved by initially computing the spectrum of the normalized Laplacian of the $k$-hop subgraph surrounding a vertex, followed by assigning a Dirac measure to it.
Later, we define a \emph{spectral distance} between two $k$-hop neighborhoods as the 
$p$-th Wasserstein distance between their corresponding probability measures.

Second, we introduce a \emph{bias} in the random walk, that with a certain probability, chooses the next vertex with least spectral distance to it. This allows our ``neighborhood-aware" walks to reach \emph{nodes of interest} much quicker and pack more such nodes in a walk of fixed length.
We refer to our biased walks as \emph{spectral-biased random walks}.

Finally, we learn embeddings for each spectral-biased walk in addition to node embeddings using a  paragraph vector model~\cite{Le2014}, such that each walk which starts at a node considers its own surrounding context within the same walk and does not share context across all the walks, in contrast to a wordvec model~\cite{mikolov2013}.
Additionally, we also add a \emph{Wasserstein regularization term} to the the objective function so that node pairs with lower spectral distance co-locate in the final embedding.

\noindent\textbf{Our contributions}
\begin{enumerate}
	\itemsep-0.5em 
	\item 
	We propose a \emph{spectral-biased random walk} that integrates neighborhood structure into the walks and makes each walk more \emph{aware} 
	of the quality of the nodes it visits.\\
	\item 
	We propose the use of paragraph vectors and a novel Wasserstein regularization term to learn embeddings for the random walks originating from a node and ensure that spectrally similar nodes are closer in the final embedding.\\
	\item 
	We evaluate our method on challenging real-world datasets for tasks such as link prediction and node classification. On many datasets, we significantly outperform our baseline methods.
	For example, our method outperforms state-of-the-art methods for two difficult datasets \emph{Power} and \emph{Road} by a margin of $6.23$ and $6.93$ in AUC, respectively.
\end{enumerate}

\section{Related Work}
Recently, several variants have been introduced to learn node embeddings for link prediction. These methods can be broadly classified  as (i) heuristic, (ii) matrix factorization, (iii) Weisfeiler-Lehman based, 
(iv) random walks based, and (v) graph neural network (GNN) methods.

Common neighbors (CN), Adamic-adar (AA)~\cite{adamic2003friends}, 
PageRank~\cite{brin2012reprint}, SimRank~\cite{jeh2002simrank}, resource allocation 
(RA)~\cite{zhou2009predicting}, preferential attachment (PA)~\cite{barabasi1999emergence}, Katz 
and resistance distance are some popular examples of heuristic methods. These methods compute 
a heuristic similarity measure between nodes to predict if they are likely to have a 
link~\cite{liben2007link}~\cite{lu2011link} between them or not. 
Heuristic methods can be further categorized into 
\emph{first-order}, \emph{second-order} and \emph{higher-order} methods based on using information from the $1$-hop, $2$-hop and $k$-hop (for $k>2$) neighborhood of 
target nodes, respectively. 
In practice, heuristic methods perform well but are based on strong assumptions 
for the likelihood of links, which can be beneficial in the case of social networks, but does not generalize well to arbitrary networks.

Similarly, a matrix factorization based approach, i.e., like spectral clustering (SC)~\cite{tang2011leveraging} also makes a strong assumption about the graph cuts being useful for classification. However, it is unsatisfactory to generalize across diverse networks.

Weisfeiler-Lehman graph kernel (WLK)~\cite{shervashidze2011weisfeiler} and 
Weisfeiler-Lehman Neural Machine (WLNM)~\cite{zhang2017weisfeiler} form an interesting class of heuristic learning methods. They are \emph{Weisfeiler-Lehman graph kernel} 
based methods, which learn 
embeddings from enclosing subgraphs in which the distance between a pair of graphs is defined as a function of the number of common rooted subtrees between both graphs. These methods have been shown to perform much better than the aforementioned traditional heuristic methods.

Other category of random walks based methods consist of DeepWalk~\cite{perozzi2014deepwalk} and Node2Vec~\cite{grover2016node2vec}, which have been proven to perform well as it pushes co-occuring nodes in a walk closer to one another in the final node embeddings. Although DeepWalk is a special case of the Node2Vec model, both of these methods produce node embeddings by feeding simple random walks to a word2vec skip-gram model~\cite{mikolov2013}.

Finally, for both link prediction and node classification tasks, recent works are mainly graph neural networks (GNNs) based architectures. VGAE~\cite{kipf2016variational}, WYS~\cite{abu2018watch}, and SEAL~\cite{zhang2018link} are some of the most recent and notable methods that fall under this category. VGAE~\cite{kipf2016variational} is a variational 
auto-encoder with a graph convolution network~\cite{kipf2017semi} as an encoder. In this, the decoder is defined by a simple inner product computed at the end. It is a node-level GNN to learn node embeddings. While WYS~\cite{abu2018watch} uses an attention model that learns context 
distribution on the power series of a transition matrix, SEAL~\cite{zhang2018link} uses a graph-level GNN and extracts enclosing subgraphs for each edge in the graph. It learns via a decaying heuristic a mapping function for link prediction. Computing subgraphs for all edges makes 
it inefficient to process large and dense graphs.
\section{Spectra of Vertex Neighborhoods}
\label{spectra_nhood}
In this section, we describe a \emph{spectral neighborhood} of an arbitrary vertex in a graph.
We start by outlining some background definitions that are relevant to our study. 
An undirected and unweighted graph is denoted by $G=(V,E)$, where $V$ is a set of 
vertices and edge-set $E$ represents a set of pairs $(u,v)$, where $u,v \in V$. Additionally, $n$ and $m$ denote the number of vertices and edges in the graph, respectively. In an undirected 
graph $(u,v) = (v,u)$. Additionally, when edge $(u,v)$ exists, we say that vertices $u$ and $v$ 
are \emph{adjacent}, or that $u$ and $v$ are \emph{neighbors}. 
The degree $d_v$ of vertex $v$ is the total number of vertices adjacent to $v$.
By convention, we disallow \emph{self-loops} and \emph{multiple edges} connecting the same
pair of vertices.
Given a vertex $v$ and a fixed integer $k > 0$, the \emph{graph neighborhood} $\mG(v,k)$ of $v$ is the subgraph induced by the $k$ closest vertices (i.e., in terms of shortest paths on $G$) that are reachable from $v$.

Now, the graph neighborhood $\mG(v,k)$ of a vertex $v$ is represented in matrix form as a \emph{normalized Laplacian matrix} $\mL^{(v)} = (l_{ij})_{i,j=1}^k \in \mathds{R}^{k^2}$. 
Given $\mL^{(v)}$, its sequence of $k$ real eigenvalues 
$(\lambda_1 (\mL^{(v)} ) \geq \dots \geq \lambda_k (\mL^{(v)} ))$ is known as the \emph{spectrum} of the neighborhood $\mL^{(v)}$ and is 
denoted by $\sigma( \mL^{(v)} ))$. We also know that all the eigenvalues in $\sigma( \mL^{(v)} ))$ lie
in an interval $ \Omega \subset \mathds{R}$. 
Let $\mu_{ \sigma( \mL^{(v)} )  }$ denote the 
probability measure on $\Omega$ that is associated to the spectrum $\sigma( \mL^{(v)} )$ and is 
defined as the Dirac mass concentrated on each eigenvalue in the spectrum. Furthermore, let 
$\mP(\Omega)$ denote the set of probability measures on $\Omega$. We now define the $p$-th
Wasserstein distance between measures, which will be used later to define our distance between 
node neighborhoods.
\begin{definition}\cite{villani2008}
	Let $p \in [1,\infty)$ and let $c: \Omega \times \Omega \rarrow{} [0,+\infty]$ be the \emph{cost function} between the probability measures $\mu,\nu \in \mP(\Omega)$. Then, the $p$-th Wasserstein distance
	between measures $\mu$ and $\nu$ is given by the formula
	\begin{equation}
	W_p( \mu, \nu ) = 
	\left(  \inf_\gamma  
	\int\displaylimits_{\Omega \times \Omega} c(x,y)^p d\gamma  \mid \gamma \in \Pi(\mu,\nu)  \right)^{\frac{1}{p}}
	\end{equation}
where $\Pi(\mu,\nu)$ is the set of \emph{transport plans}, i.e., the set of all joint probabilities defined 
on $\Omega \times \Omega$ with marginals $\mu$ and $\nu$.
\end{definition}
We now define the \emph{spectral distance} between two vertices $u$ and $v$, along with their 
respective neighborhoods $\mL^{(u)}$ and $\mL^{(v)} $, as
\begin{equation}
\label{eq:2}
W^p( u, v ) := W_p( \mu_{ \sigma( \mL^{(u)} )  }, \mu_{ \sigma( \mL^{(v)} )  }  )
\end{equation}

\section{Random Walks on Vertex Neighborhoods}
\subsection{Simple random walk between vertices} 
A simple random walk on $G$ begins with the choice of an initial 
vertex $v_0$ chosen from an initial probability distribution on $V$ at time $t_0$. For each time $t \geq 0$, the next vertex to move to is chosen \emph{uniformly at random} from the current 
vertex's $1$-hop neighbors.
Hence, the probability of transition $p_{ij}$ from vertex $i$
to its $1$-hop neighbor $j$ is $1/d_i$ and $0$ otherwise. This stochastic process is a \emph{finite Markov chain} and the non-negative matrix $P= (p_{ij})_{i,j=1}^n \in \mathds{R}^{n \times n}$ is its corresponding \emph{transition matrix}. We will focus on ergodic finite Markov chains with a \emph{stationary distribution} 
$\pi^T = (\pi_1, \dots, \pi_n)$, i.e., $\pi^T P = \pi^T$ and $\sum_{i=1}^{n} \pi_i = 1$.  
Let $\{ X_t \}$ denote a Markov chain (random walk) with state space $V$. Then, the \emph{hitting time}
for a random walk from vertex $i$ to $j$ is given by 
$T_{ij} = \inf \{ t: X_t = j \mid X_0 = i  \}$
and the \emph{expected hitting time} is $\E[T_{ij} ]$. In other words, hitting time $T_{ij}$ is the first 
time $j$ is reached from $i$ in $\{ X_t \}$.
By the \emph{convergence theorem}~\cite{AldousBook2002}, we know that the transition matrix $P$ satisfies 
$\lim\limits_{n \rarrow{} \infty} P^n = P_\infty$, where matrix $P_\infty$ has all its rows equal to $\pi$.

\subsection{Spectral-biased random walks} 
We introduce a \emph{bias} based on the \emph{spectral distance} 
between vertices (as shown in Equation~\ref{eq:2}) in our random walks.
When moving from a vertex $v$ to an adjacent vertex $v'$ in the $1$-hop neighborhood $\mN(v)$ of vertex $v$,  
vertices in $\mN(v)$ which are most \emph{structurally similar} to $v$ are \emph{favored}. 
The most structurally similar vertex to $v$ is given by 
\begin{equation}
\min_{ v' \in \mN(v) } W^p(v,v')
\end{equation}
Then, our \emph{spectral-biased walk} is a random walk where each of its step is preceded by the
following decision. Starting at vertex $i$, the walk transitions with probability $1-\epsilon$ to an 
adjacent vertex $j$ in $\mN(v)$ uniformly at random, and with probability $\epsilon$, 
the walk transitions to the next vertex with probability $w_{ij}$ given in the bias matrix, whose detailed construction is explained later. 
Informally, our walk can be likened to 
flipping a biased coin with probabilities $1-\epsilon$ and $\epsilon$, prior to each move, to decide whether to perform a simple random walk or choose one of $k$ structurally similar nodes from the neighborhood.
Thus, our new spectral-biased transition matrix can be written more succinctly as
\begin{equation}
T = (1-\epsilon) P + \epsilon W
\end{equation}
where $P$ is the original transition matrix for the simple random walk and $W$ contains the biased transition probabilities we introduce to move towards a structurally similar vertex.
\begin{figure*}[tbp]
	\begin{subfigure}{.33\textwidth}
		\includegraphics[width=\linewidth]{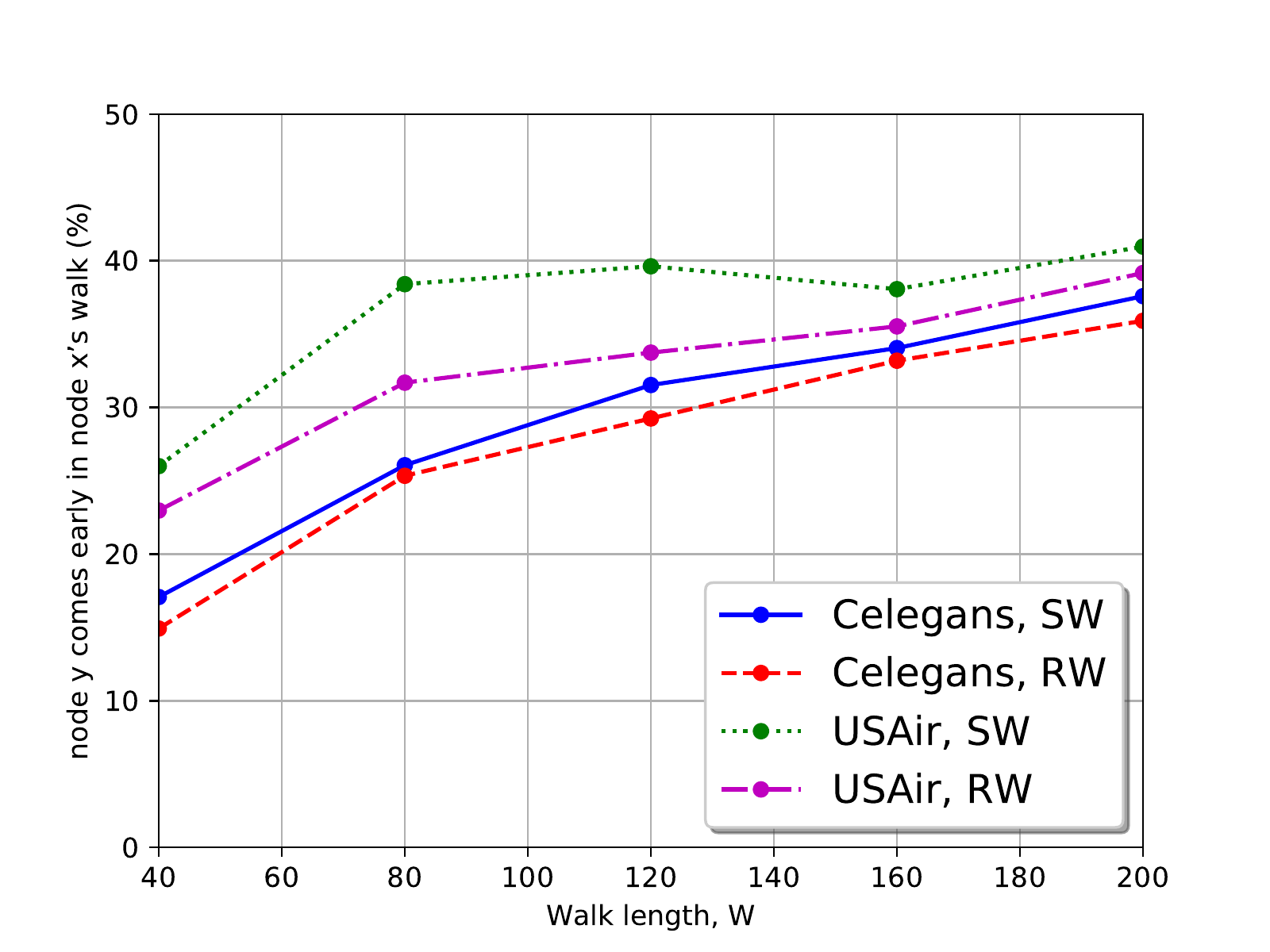}
		\caption{}
		\label{fig:sub2}
	\end{subfigure}
	\begin{subfigure}{.33\textwidth}
		\includegraphics[width=\linewidth]{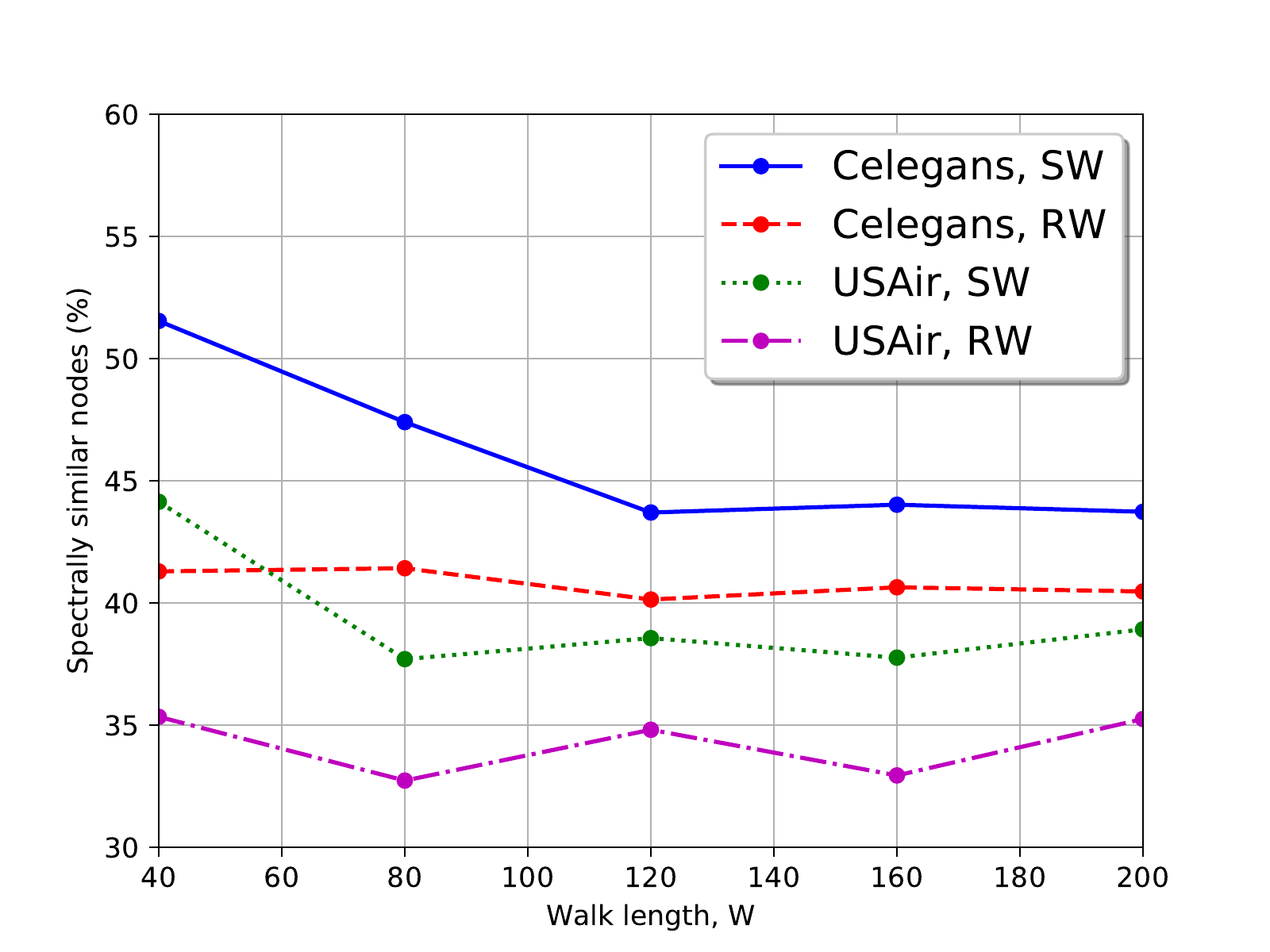}
		\caption{}
		\label{fig:sub3}
	\end{subfigure}
	\begin{subfigure}{.33\textwidth}
		\includegraphics[width=0.9\linewidth]{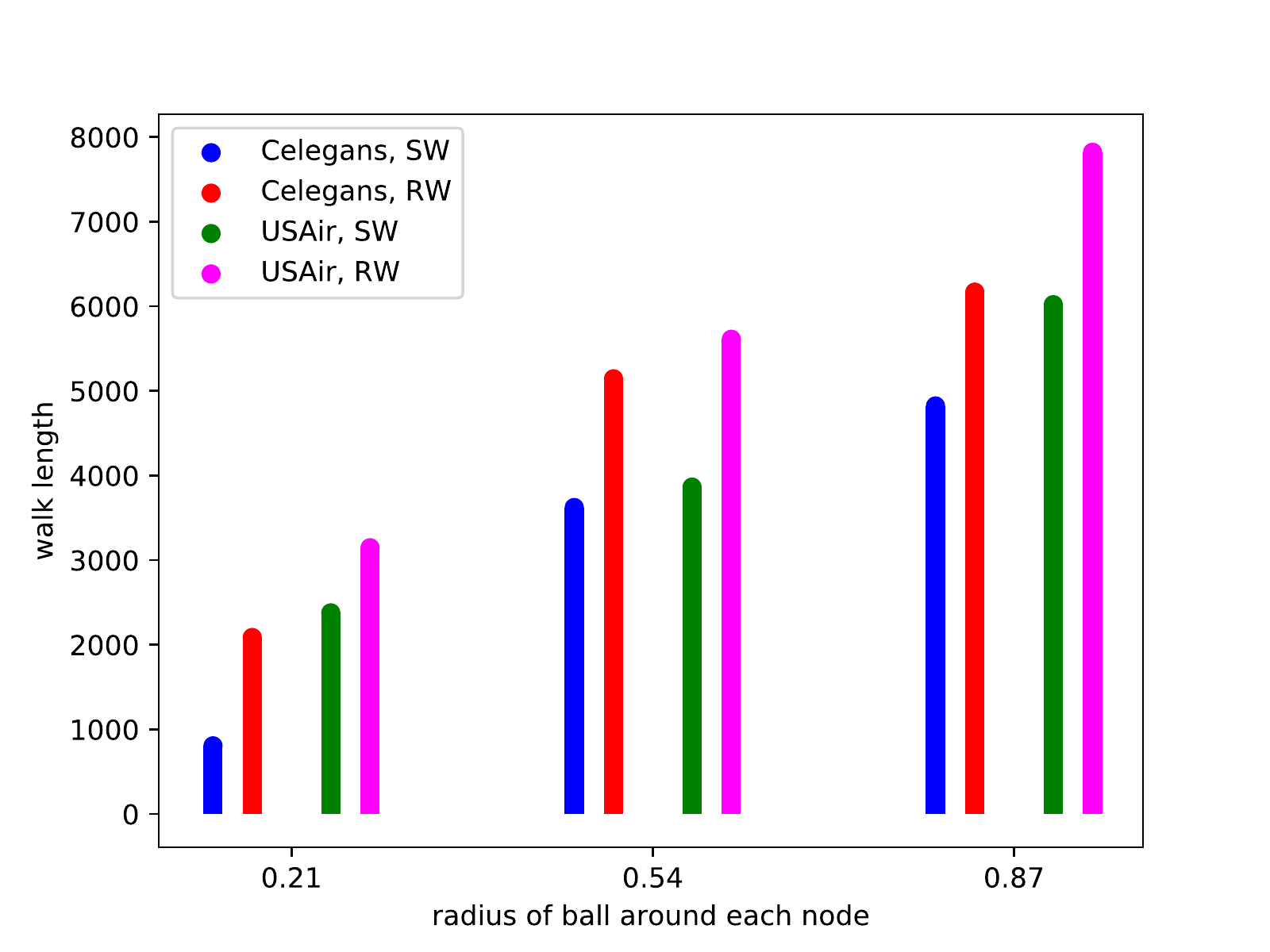}
		\caption{}
		\label{fig:sub1}
	\end{subfigure}
	\begin{subfigure}{.33\textwidth}
		\includegraphics[width=\linewidth]{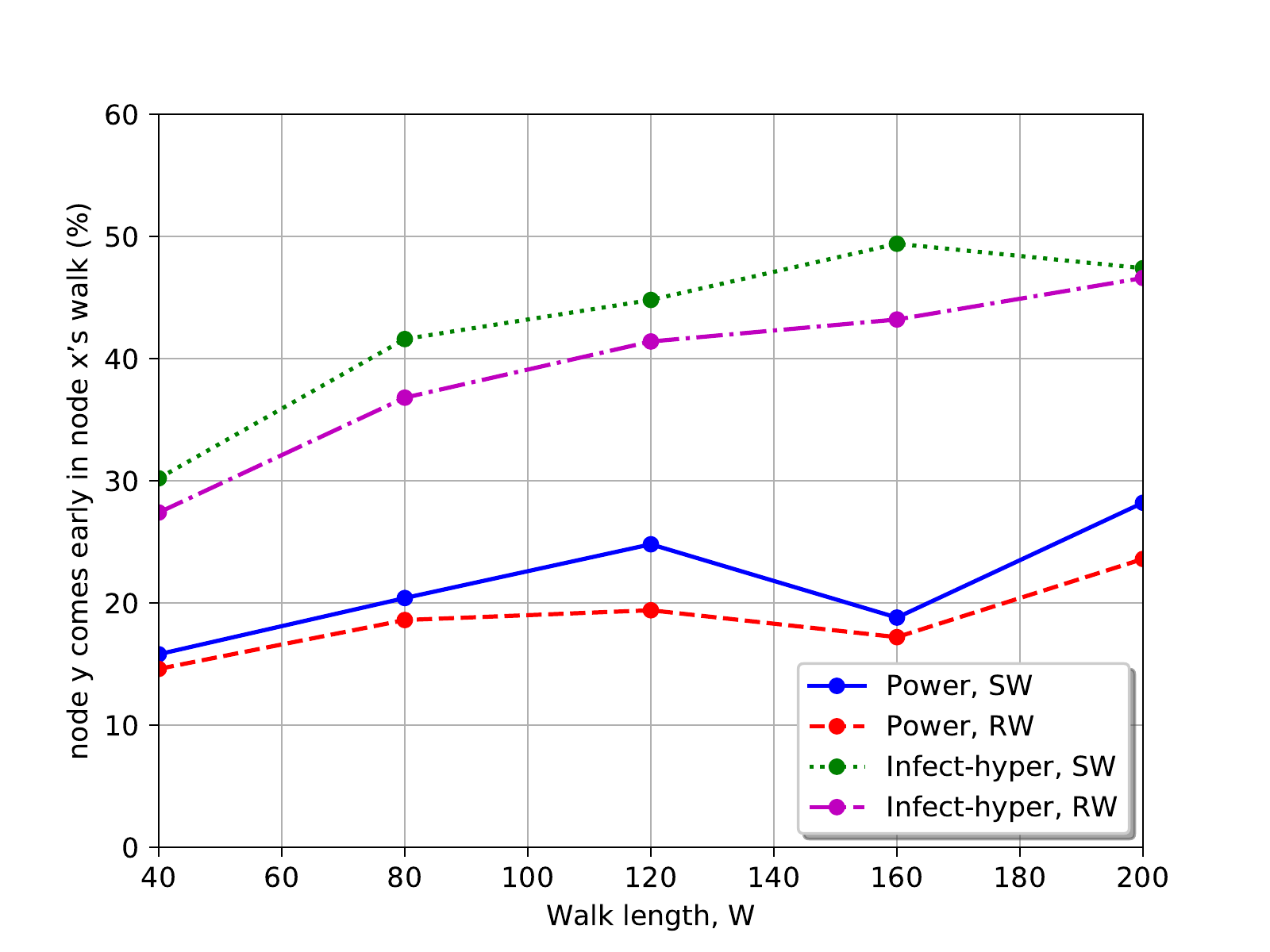}
		\caption{}
		\label{fig:sub5}
	\end{subfigure}
	\begin{subfigure}{.33\textwidth}
		\includegraphics[width=\linewidth]{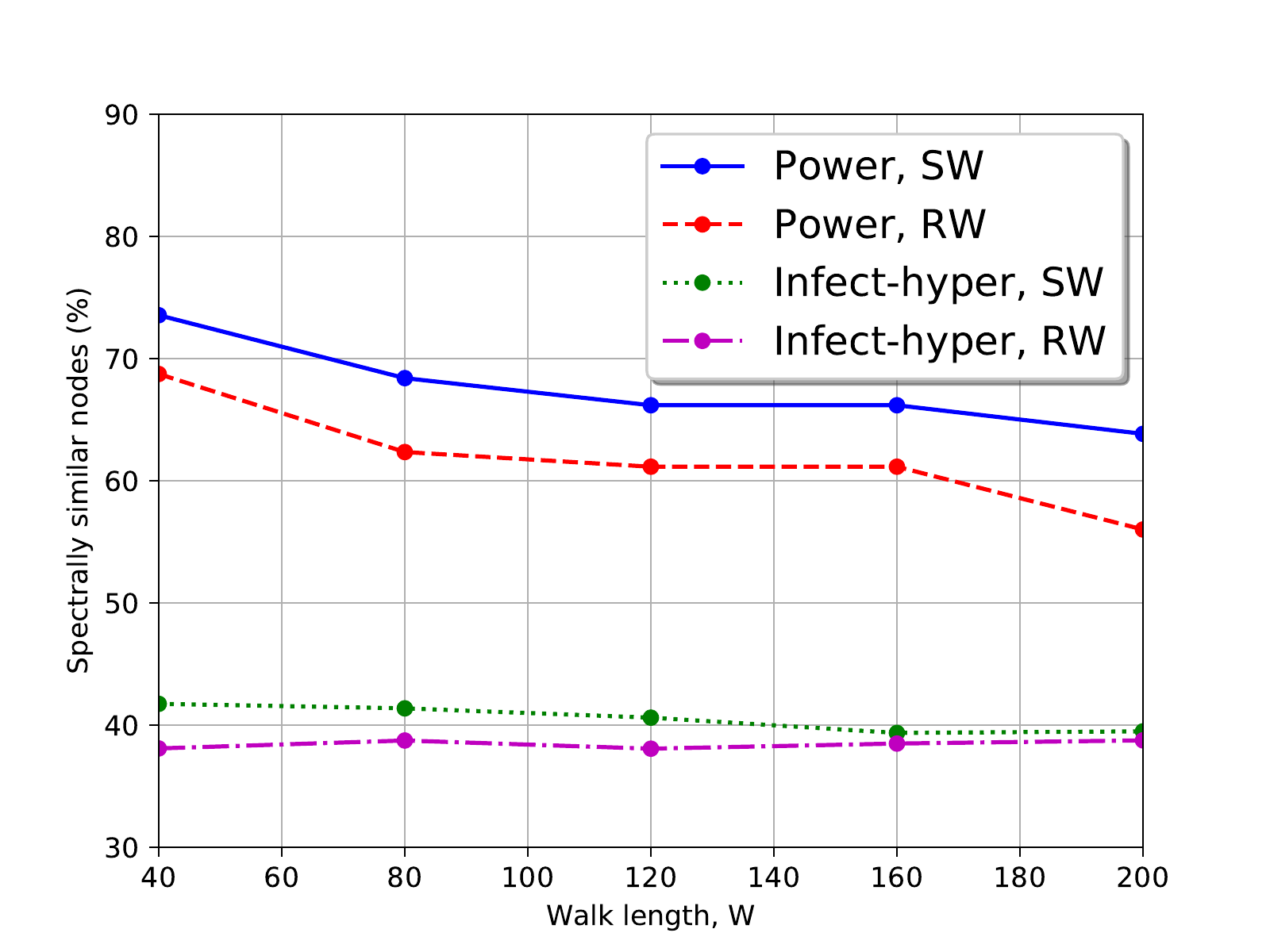}
		\caption{}
		\label{fig:sub6}
	\end{subfigure}
	\begin{subfigure}{.33\textwidth}
		\includegraphics[width=\linewidth]{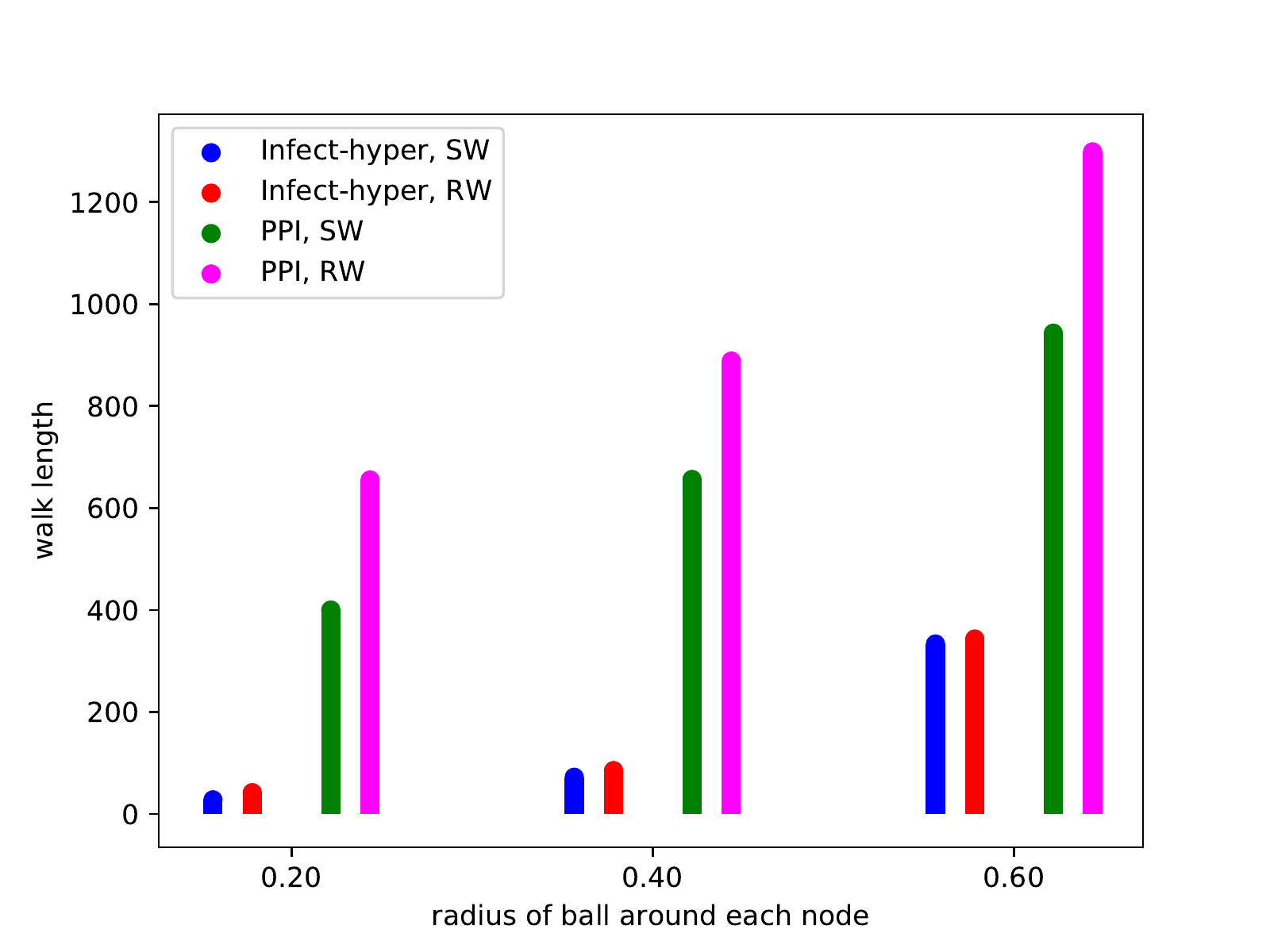}
		\caption{}
		\label{fig:sub4}
	\end{subfigure}
	\caption{(a) and (d) Average ranking of target nodes encountered by simple 
		random walk (RW) and our spectral-biased walk (SW). (b) and (e) Percentage of spectrally-similar nodes packed in walks of varying length for RW versus. SW. (c) and (f) 
		Walk lengths to cover entire ball of vertices on Celegans, USAir, Infect-hyper and PPI for our spectral biased walk (SW) and simple random walk (RW). }
	\label{fig:spectral_walk}
\end{figure*}
 \subsection{Spectral bias matrix construction}
 It is well known that the spectral decomposition of a symmetric stochastic matrix produces \emph{real eigenvalues} in the interval $[-1,1]$. In order to build a biased transition matrix $W$ which allows the spectral-biased walk to take control with probability $\epsilon$ and choose among $k$ nearest neighboring vertices with respect to the spectral distance between them, we must construct this bias matrix in a special manner. Namely, it should represent a \emph{reversible} Markov chain, so that it can be ``symmetrized". For brevity, we omit a detailed background necessary to understand \emph{symmetric transformations}, but we refer the reader to~\cite{norris1998}. 
 A Markov chain is said to be \emph{reversible}~\cite{LevinPeresWilmer2006}, when it satisfies the \emph{detailed balance condition} 
 $\pi_i p_{ij} = \pi_j p_{ji}$, i.e., on every time interval of the stochastic process the distribution of the process is the same when it is run forward as when it is run backward.
 
 Recall, the $1$-hop neighborhood of vertex $i$ is denoted by $\mN(i)$. Additionally, we define 
 $\mN_k(i) \subseteq \mN(i)$ to be the $k$\emph{-closest} vertices in spectral distance 
 to $i$ among $\mN(i)$.
 
 We then define a \emph{symmetric  $k$-closest neighbor set} $\mS_k(i)$ as a union of all the members of $\mN_k(i)$ and those vertices $j \in \mN(i) \setminus \mN_k(i) $, who have vertex $i$ in $\mN_k(j)$. More formally,
 \begin{equation}
 \mS_k(i) := \mN_k(i) \cup \left\{   \bigcup_{j \in \mN(i) \setminus \mN_k(i)  }    \mathds{1}_{\mN_k(j)} (i)            \right\} 
 \end{equation}
where the indicator function $\mathds{1}_A(x) := 1$, if $x \in A$ or $0$ if $x \notin A$.  

In accordance to \emph{property} $7.1.1$ in~\cite{Qian2009}, we construct a transition matrix as follows to form a \emph{reversible Markov chain} which satisfies the detailed balance condition and hence is symmetrizable. 
Our bias matrix $W = (w_{ij})_{i,j=1}^n$ is a stochastic transition probability matrix in $\R^{n^2}$, whose elements are given by
\begin{align}
\label{eq:bias}
w_{ij} = \left \{    1 - \frac{W^p(i,j)}{\sum_{m \in S_k(i)} W^p(i,m)   }  \right \} 
\end{align}
The rows of the spectral bias matrix $W$ in Equation~\ref{eq:bias} are scaled appropriately to convert it into a transition matrix. 

\subsection{Time complexity of our spectral walk}
Given $n$ nodes in a graph, we first pre-compute the spectra of every vertex's neighboring subgraph (represented as a normalized Laplacian). This spectral computation per vertex includes spectral decomposition of the Laplacian around each vertex, which has a time complexity of $O(k^{2.376})$ (where, $k$ is the size of each vertex neighborhood, typically of $O(10)$, which is very 
fast to compute) using the \emph{Coppersmith and Winograd algorithm} for matrix multiplication, which is the most dominant cost in decomposition.
This amounts to a total pre-computation time complexity of $O(nk^{2.376})$.

In the worst case, a spectral-biased walk of length $l$ will be biased at each step and hence would compute the spectral distance among its $k$ neighbors at each step (i.e., a total of $kl$ times).  
The Wasserstein distance between the spectra of the neighborhoods has an empirical time complexity of $O(d^2)$, where $d$ is the order of the histogram of spectra $\sigma( \mL^{(v)} )$. 
Thus the time-complexity of our online spectral-biased walk is $O(kld^2)$.
Although, in practice, we use the Python OT library based on entropic regularized OT, which uses the Sinkhorn algorithm on a GPU and thus computing Wasserstein distances are extremely fast and easy.



\subsection{Empirical analysis of expected hitting time and cover time of spectrally similar vertices}

In this section, we empirically study the quality of the random walks produced by our spectral-bias random walk method. In order to accomplish this, we start with a given vertex $v$ and measure the walk quality under two popular quality metrics associated with random walks, namely their \emph{expected hitting time} and \emph{cover time} of nodes with structurally similar neighborhoods to that of node $v$. It is important to note here that the consequence of packing more nodes of interest in each random walk, boosts the quality of training samples (i.e., walks setup as \emph{sentences}) in our neural language model that is described later in Section~\ref{sec:nn}.

%

\textbf{Expected hitting time:} To study the expected hitting times of our spectral-biased and simple random walks, 
we first randomly sampled $1000$ ordered vertex pairs $(s,t)$ with structurally similar neighborhoods, where $s$ and $t$, denoted the 
\emph{start} and \emph{target} vertices, respectively. 
Next, we considered all the random walks (both spectral-biased and simple) initiated from the start vertex $s$ and ranked the appearance of the target vertex $t$ in a fixed length walk, for both the types of walks. Our ranking results where averaged over all the walks and $(s,t)$ pairs considered.
In our experiments on real-world datasets (shown in Figures~\ref{fig:sub2} and~\ref{fig:sub5}), we found the target vertex $t$ to appear earlier in our spectral-biased walks, i.e., we had a \emph{lower expected hitting time} from $s$ to $t$.  

Furthermore, we also studied the \emph{packing density} of spectrally similar nodes in fixed-length walks generated by both the spectral-bias and simple random walk methods. Figures~\ref{fig:sub3} and~\ref{fig:sub6}, clearly show that our spectral-biased walk packs a higher number of spectrally similar nodes.

\textbf{Cover time:} 
After having empirically studied the spectral-biased walk's expected hitting time, it naturally leads to study the \emph{cover time} of our walk, which is the first time when all vertices that are spectrally similar to a start vertex have been visited. 

We begin by defining a \emph{Wassertein ball} around an arbitrary vertex $v$ that encompasses the set of  vertices whose \emph{spectral distance} from $v$ is less than a constant $c$. 
\begin{definition}
	A Waserstein ball of radius $c$ centered at vertex $v$, denoted by $B_w(v;c)$, is defined as
	\begin{equation}
	B_w(v;c) := \{  u \in V \mid W^p(u,v) \leq c  \}
	\end{equation}
\end{definition}
Given a start vertex $s$, a user-defined fixed constant $c$,  and its surrounding Wasserstein ball $B_w(s;c)$, we found that our spectral-bias walk covers all spectrally similar vertices in the ball with much \emph{shorter} walks than simple random walks, as is shown in Figures~\ref{fig:sub1} and~\ref{fig:sub4}.

\begin{table*}[tbp]
	\caption{Link prediction results (AUC). "-" for incomplete execution due to either \emph{out of memory errors} or runtime exceeding $20$ hours. Bold indicate best and underline indicate second best results.}
	\centering
	\footnotesize
	\setlength\tabcolsep{3.5pt}
	\begin{tabular}{llllllll}
		\toprule
		Algorithms & Node2Vec & VGAE & WLK & WLNM & SEAL & WYS & \textbf{Our Method} \\
		\hline
		
		Power & 78.37 $\pm$ 0.23 & 77.77 $\pm$ 0.95 & - & - & 74.69 $\pm$ 0.21 & \underline{89.37 $\pm$ 0.21} &  \textbf{95.60 $\pm$ 0.25}  \\
		Celegans & 69.85 $\pm$ 0.89 & 74.16 $\pm$ 0.78 & 73.27 $\pm$ 0.41 & 70.64 $\pm$ 0.57 & \underline{85.53 $\pm$ 0.15} & 74.97 $\pm$ 0.19 & \textbf{87.36 $\pm$ 0.10}  \\
		USAir & 84.90 $\pm$ 0.41 & 93.18 $\pm$ 1.46 & 87. 98 $\pm$ 0.71 & 87.01 $\pm$ 0.42 & \underline{96.9 $\pm$ 0.37} & 94.01 $\pm$ 0.23 & \textbf{97.40 $\pm$ 0.21} \\
		Road-Euro & 50.35 $\pm$ 1.05  & 68.94 $\pm$ 5.23 & 61.17 $\pm$ 0.28 & 65.95 $\pm$ 0.33 & 60.89 $\pm$ 0.22 & \underline{80.42 $\pm$ 0.11} & \textbf{87.35 $\pm$ 0.33}\\ 
		Road-Minnesota & 67.12 $\pm$ 0.63 & 67.36 $\pm$ 2.33 & 75.15 $\pm$ 0.16 & 74.91 $\pm$ 0.19 & \underline{86.92 $\pm$ 0.52} & 75.33 $\pm$ 2.77 & \textbf{91.16 $\pm$ 0.15} \\
		Bio-SC-GT & 88.39 $\pm$ 0.79 & 86.76 $\pm$ 1.41 & - & - & \textbf{97.26 $\pm$ 0.13} & 87.72 $\pm$ 0.47 & \underline{97.16 $\pm$ 0.32}\\
		Infect-hyper & 66.66 $\pm$ 0.51 & 80.89 $\pm$ 0.21 & 65.39 $\pm$ 0.39 & 67.68 $\pm$ 0.41 & \underline{81.94 $\pm$ 0.11} & 78.42 $\pm$ 0.15 & \textbf{85.25 $\pm$ 0.24} \\
		PPI & 71.51 $\pm$ 0.09 & \underline{88.19 $\pm$ 0.11} & - & - & - & 84.12$\pm$ 1.27 & \textbf{91.16 $\pm$ 0.30}\\
		Facebook & 96.33 $\pm$ 0.11 & - & - & - & - & \underline{98.71 $\pm$ 0.14} & \textbf{99.14 $\pm$ 0.05}\\
		HepTh & 88.18 $\pm$ 0.21 & 90.78 $\pm$ 1.15 & - & - & \textbf{97.85 $\pm$ 0.39} & 93.63 $\pm$ 2.36 & \underline{97.40 $\pm$ 0.25} \\
		\bottomrule
	\end{tabular}
	\label{tab:linkpred}
\end{table*}

\section{Our Neural Language Model with Wasserstein Regularization}
\label{sec:nn}
Our approach of learning node embeddings is to use a shallow neural network. This network takes 
spectral-biased walks as input and predicts either the node labels for node classification or the likelihood of an edge / link between a pair of nodes for the link prediction task. 

We leverage the similarity of learning paragraph vectors in a document from NLP to learn our spectral-biased walk embeddings. 
In order to draw analogies to NLP, we consider a \emph{vertex} as a \emph{word}, a \emph{walk} as a \emph{paragraph / sentence}, 
and the entire \emph{graph} as a \emph{document}. Two walks are said to \emph{co-occur} when they originate from the same node. Originating from each node $v \in V$, we generate $K$ co-occurring 
spectral-biased random walks $\mW^{(v)} = ( \mW_1^{(v)}, \dots ,   \mW_K^{(v)}  )$, each of fixed length $T$. A family of all $\mW^{(v)}$ for all $v \in V$ is analogous to a collection of paragraphs in a document.

In our framework, each vertex $v$ is mapped to a unique word vector $w$, represented as a column in a matrix $W$. Similarly, each biased walk $w$ is mapped to a unique paragraph vector $p$ stored as a column in a matrix $P$. 
Given a spectral-biased walk as a \emph{sequence of words} $w_1, w_2, \dots, w_T$, our objective is to minimize the following cross-entropy loss
\begin{equation}
L_{par} = - \frac{1}{T} \sum_{t=c}^{T-c} \log p( w_t \mid w_{t-c}, \dots , w_{t+c})
\end{equation}
As shown in~\cite{Le2014}, the probability is typically given by the \emph{softmax} function 
\begin{equation}
p( w_t \mid w_{t-c}, \dots , w_{t+c}) = \frac{e^{y_{w_t}}  }{   \sum_{i} e^{y_{w_i }}}
\end{equation}
Each $y_{w_i}$ is the unnormalized log probability for $w_i$, given as
$y_{w_t} = b + Uh( w_{t-c}, \dots, w_{t+c} ; P,W   )$, 
where $U,b$ are softmax parameters, and $h$ is constructed from $W$ and $P$.
A paragraph vector can be imagined as a word vector that is cognizant of the context information encoded in its surrounding paragraph, while a normal word vector \emph{averages} this information across all paragraphs in the document. For each node $v \in V$, we apply $1$d-convolution to all the paragraphs / walks in $\mW^{(v)}$, to get a unique vector $x_v$.

Our goal is to learn node embeddings which best preserve the underlying graph structure along with clustering structurally similar nodes in feature space. 
With this goal in mind and inspired by the work of Mu et. al.~\cite{mu2018} on \emph{negative skip-gram sampling with quadratic regularization}, we construct the following loss function with a Wasserstein regularization term
\begin{equation}
\label{eq:overall_loss}
L_{ov} = L_{par} + L_{class} +  \underbrace{  \gamma W^2( \sigma^{(s)}(x_v) , \sigma^{(s)}(y_v))}_{\text{$2$-Wasserstein regularizer}}
\end{equation}
Here, $x_v$ is the node embedding learned from the paragraph vector model and $y_v$ is the 
$1$d-convolution of node $v$'s $1$-hop neighbor embeddings. $L_{class}$ represents a 
task-dependent \emph{classifier loss} which is set to \emph{mean-square error} (MSE) for \emph{link prediction} and \emph{cross-entropy loss} for \emph{node classification}. We convert the node embedding $x_v$ and its combined $1$-hop neighborhood embedding $y_v$ into \emph{probability distributions} via the softmax function, denoted by $\sigma^{(s)}$ in Equation~\ref{eq:overall_loss}.

Our regularization term is the $2$-Wasserstein distance between the two probability distributions, 
where $\gamma$ is the regularization parameter. This regularizer penalizes neighboring nodes whose neighborhoods do not bear structural similarity with the neighborhood of the node in question. Finally, the overall loss $L_{ov}$ is minimized across all nodes in $G$ to arrive at final node embeddings.

\section{Experimental Results}
We conduct exhaustive experiments to evaluate our spectral-biased walk 
method\footnote{\href{https://github.com/charusharma1991/LinkPred}{Our Method}}. Network datasets were sourced from \href{https://snap.stanford.edu/data/}{SNAP} and \href{http://networkrepository.com/}{Network Repository}. 
We picked ten datasets for \emph{link prediction} experiments, as can be seen in Table~\ref{tab:linkpred}, and three datasets (i.e., Cora, Citeseer, and Pubmed) for \emph{node classification} evaluation. 
The dataset statistics are outlined in more detail in Section~\ref{sec:datasets}.
We performed experiments 
by making $90\%-10\%$ train-test splits on both \emph{positive} (existing edges) and \emph{negative} (non-existent edges) samples from the graphs, following the split ratio outlined in SEAL~\cite{zhang2018link}.
We borrow notation from WYS~\cite{abu2018watch} and similarly denote our set of edges for training and testing as $E_{train}$ and $E_{test}$, respectively.

\begin{table}[h]
	\caption{Datasets for link prediction and node classification tasks.}
	\centering
	\scriptsize
	\setlength\tabcolsep{3.5pt}
	\begin{tabular}{llllllll}
		\toprule
		Datasets  & Nodes & Edges & Mean Degree & Median Degree \\
		\hline
		
		Power &  4941 & 6594 & 2.66  & 4  \\
		Celegans & 297 & 2148 & 14.46 & 24 \\
		USAir &  332 & 2126 & 12.8 & 10 \\
		Road-Euro & 1174 & 1417 & 2.41 & 4 \\
		Road-Minnesota & 2642 & 3303 & 2.5 & 4 \\
		Bio-SC-GT & 1716 & 33987 & 39.61 &  41\\
		Infect-hyper & 113 & 2196 & 38.86 & 74 \\
		PPI & 3852 & 37841 & 19.64 & 18 \\
		Facebook & 4039 & 88234 & 43.69 & 50 \\
		HepTh & 8637 & 24805 & 5.74 & 6 \\
		\hline
		Cora & 2708 & 5278 & 3.89 & 6  \\
		Pubmed &19717 & 44324 & 4.49 &  4\\
		Citeseer & 3327 & 4732 & 2.77 & 4\\
		\bottomrule
	\end{tabular}
	\label{tab:datasets}
\end{table}

\subsection{Datasets}\label{sec:datasets}
We used ten datasets for link prediction experiments and three datasets for node classification experiments. Datasets for both the experiments are described with their statistics in Table~\ref{tab:datasets}. Power~\cite{sen2008collective} is the electrical power network 
of US grid, Celegans~\cite{sen2008collective} is the neural network of the nematode worm C.elegans, USAir~\cite{nr} is an infrastructure network of US Airlines, Road-Euro and Road-Minnesota~\cite{nr} are road networks (sparse), Bio-SC-GT~\cite{nr} is a biological network of WormNet, 
Infect-hyper~\cite{nr} is a proximity network, PPI~\cite{stark2006biogrid} is a network of protein-protein interactions, HepTh is a citation network and Facebook is a social network. Cora, Citeseer and Pubmed datasets for node classification are citation networks 
of publications~\cite{sen2008collective}.

\subsection{Training}
\begin{figure*}[tbp]
	\centering
	\begin{subfigure}[b]{0.25\textwidth}
		\includegraphics[width=\linewidth]{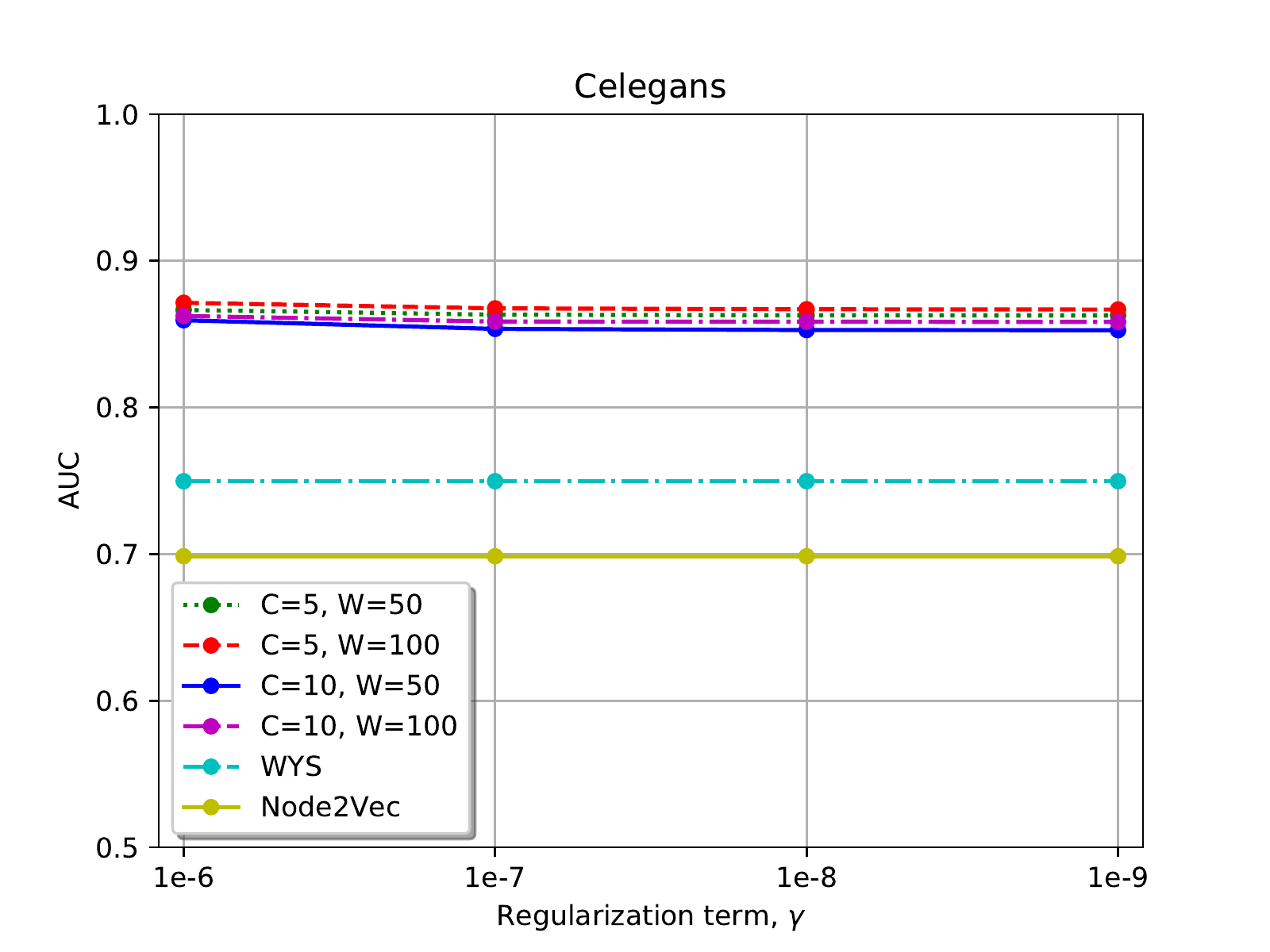}
	\end{subfigure}%
	\begin{subfigure}[b]{0.25\textwidth}
		\includegraphics[width=\linewidth]{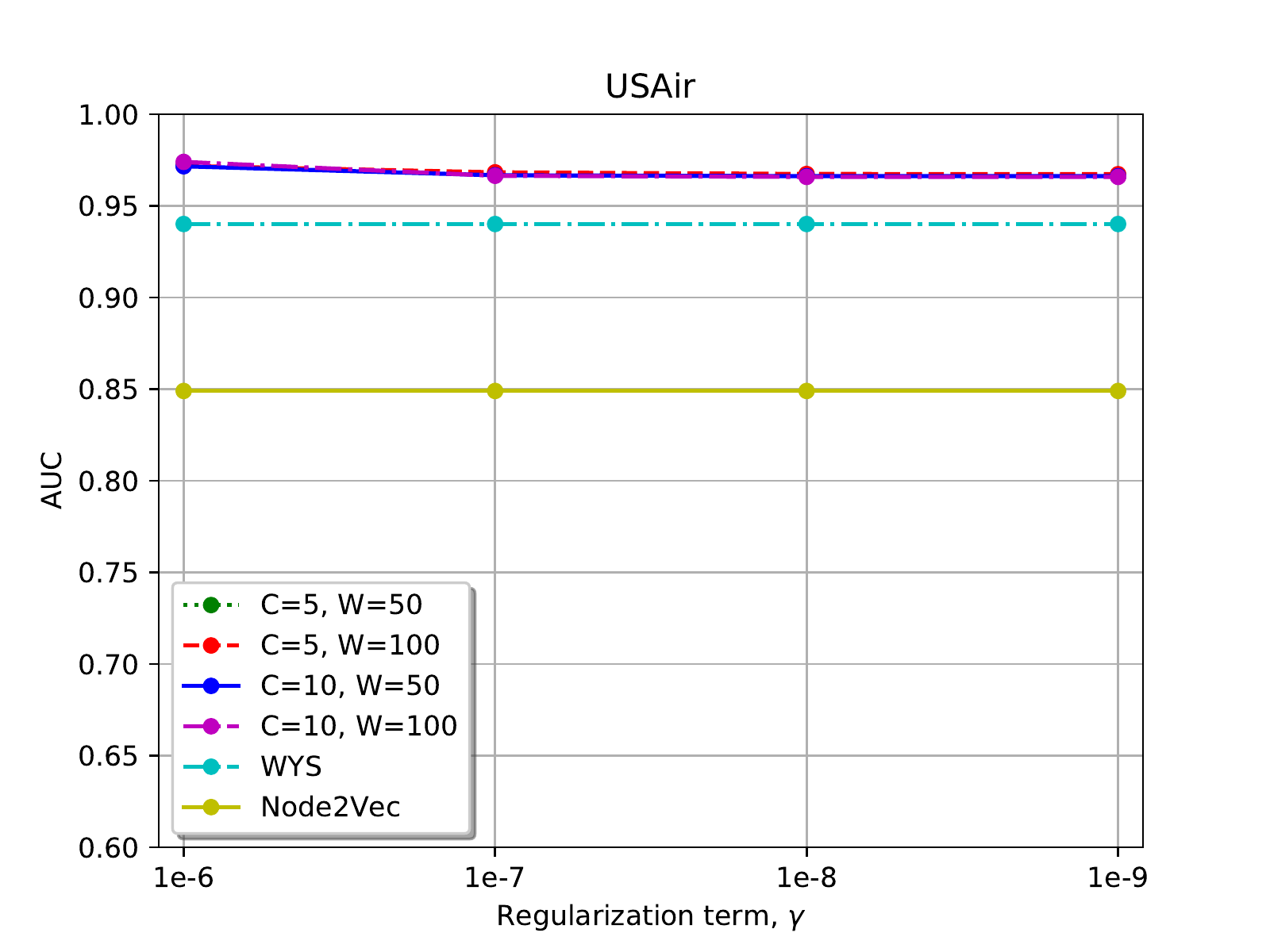}
	\end{subfigure}%
	\begin{subfigure}[b]{0.25\textwidth}
		\includegraphics[width=\linewidth]{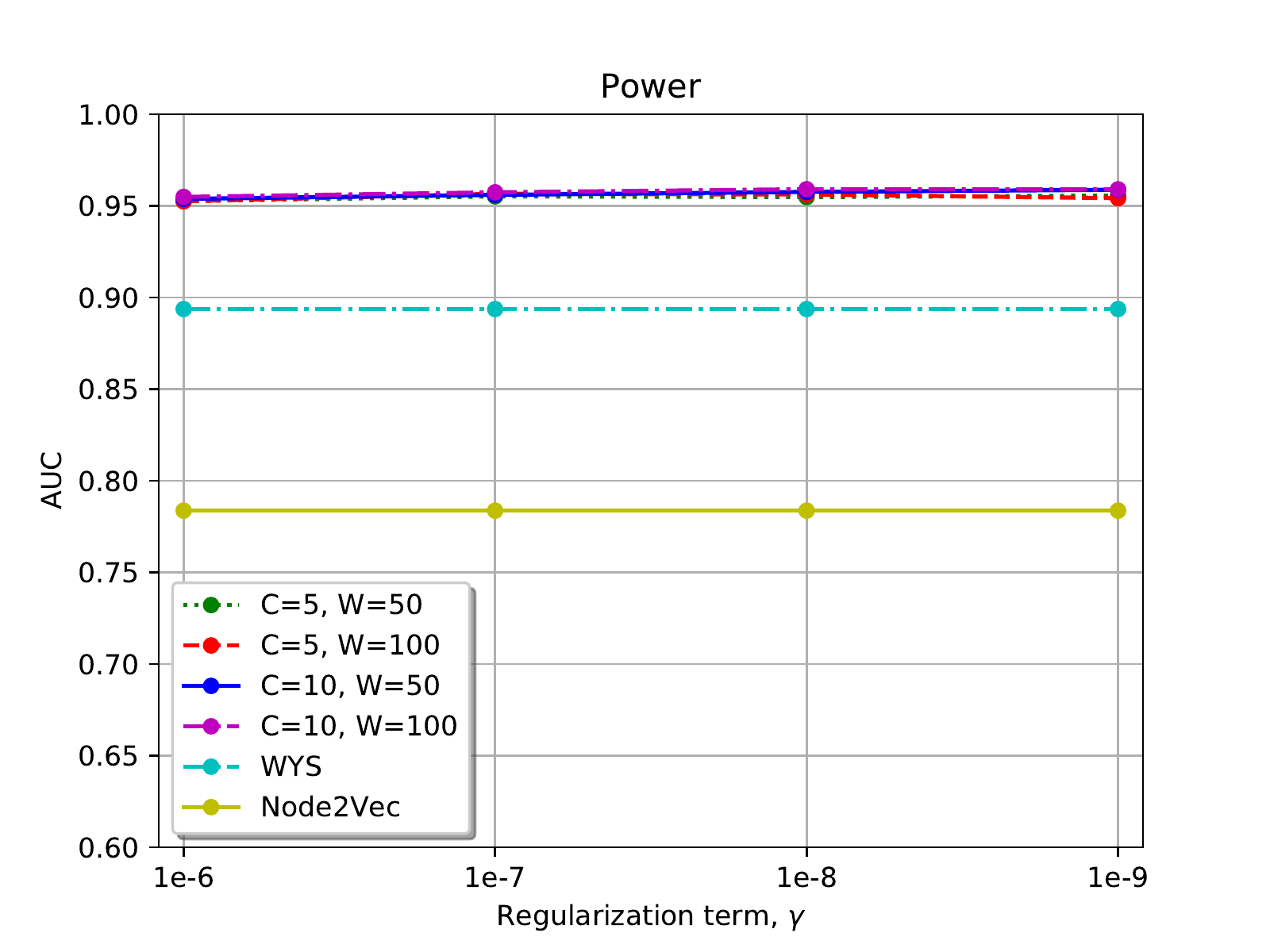}
	\end{subfigure}%
	\begin{subfigure}[b]{0.25\textwidth}
		\includegraphics[width=\linewidth]{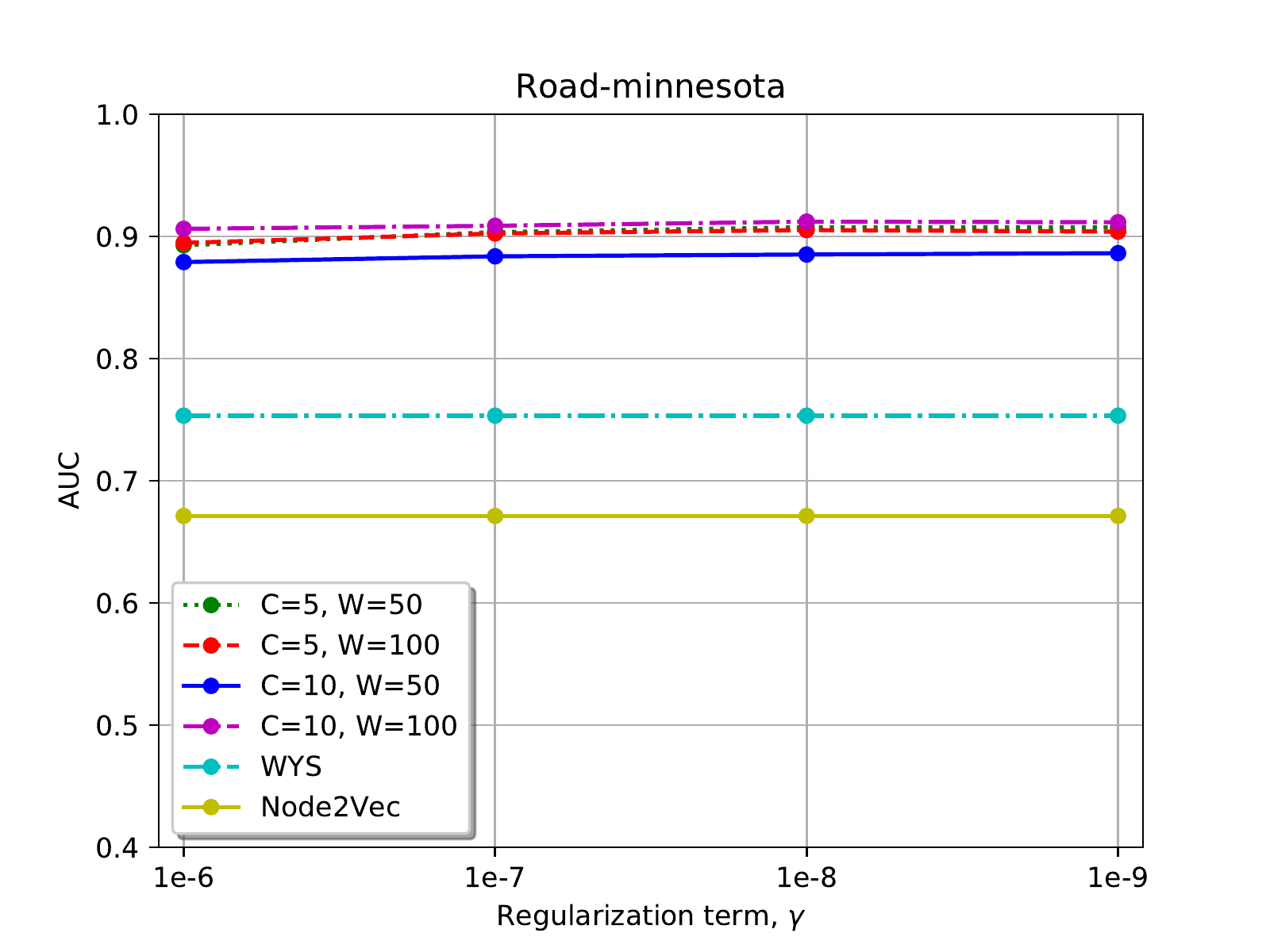}
	\end{subfigure}%
	\caption{Sensitivity of window size, $C$ and walk length, $W$ with respect to regularization term, $\gamma$ is measured in AUC for four datasets of link prediction.}
	\label{fig:senstivity}
\end{figure*}

We now turn our attention to a two-step procedure for training. First, we construct a $2$-hop neighborhood around each node for spectra computation. Probability $p$ is set to $0.6$, walk length $W=100$ with $50$ walks per node in step one of the spectral-biased walk 
generation. Second, the context window size $C=10$ and 
regularization term $\gamma$ ranges from $1e-6$ to $1e-8$ for all the results provided in Table~\ref{tab:linkpred}. The model for link prediction task to compute final AUC is trained for $100$ to $200$ epochs depending on the dataset. The 
dimension of node embeddings is set to $128$ for all the cases and a model is learned with a single-layer neural network as a classifier. We also analyze
sensitivity of hyper-parameters in Figure~\ref{fig:senstivity} to show the robustness of our
algorithm. Along with sensitivity, we also discuss how probability $p$ affects the quality of our walk in Figure~\ref{fig:p}.
%
%
%
%

\subsection{Baselines}
Our baselines are based on graph kernels (WLK~\cite{shervashidze2011weisfeiler}), GNNs (WYS~\cite{abu2018watch}, SEAL~\cite{zhang2018link}, VGAE~\cite{kipf2016variational}, and WLNM~\cite{zhang2017weisfeiler}) and random walks (Node2Vec~\cite{grover2016node2vec}). We use available codes for all the methods and evaluate the methods by computing the \emph{area under curve} (AUC). WYS~\cite{abu2018watch} learns context distribution by using an attention model on the power series of a transition matrix\footnote{\href{https://github.com/google-research/google-research/tree/master/graph_embedding/watch_your_step}{WYS}}.
On the other hand, SEAL~\cite{zhang2018link} extracts a local subgraph around each link and learns via a \emph{decaying heuristic} a mapping function to predict links\footnote{\href{https://github.com/muhanzhang/SEAL/tree/master/Python}{SEAL}}.
VGAE~\cite{kipf2016variational} is a graph based variational auto-encoder (VAE) with a graph convolutional network (GCN)~\cite{kipf2017semi} as an \emph{encoder} and simple inner product computed at the decoder side\footnote{\href{https://github.com/tkipf/gae}{VGAE}}.
A graph kernel based approach is the \emph{Weisfeiler-Lehman graph kernel} (WLK)~\cite{shervashidze2011weisfeiler}, where the distance between a pair of graphs is defined as a function of the number of common rooted subtrees between both graphs.
\emph{Weisfeiler-Lehman Neural Machine } (WLNM)~\cite{zhang2017weisfeiler} is neural network 
training model based on the WLK 
algorithm\footnote{\href{https://github.com/muhanzhang/LinkPrediction}{WLNM}}.
Node2Vec~\cite{grover2016node2vec} produces node embeddings based on generated simple random 
walks that are fed to a word2vec skip-gram model for 
training\footnote{\href{https://github.com/aditya-grover/node2vec}{Node2Vec}}.

\subsection{Link prediction}
This task entails removing links / edges from the graph and then measuring the ability of an embedding algorithm to infer such missing links.
We pick an equal number of \emph{existing edges} (``positive" samples) $E^+_{train}$ and \emph{non-existent edges} (``negative" samples) $E^-_{train}$ from the training split $E_{train}$ and 
similarly pick positive $E^+_{test}$ and negative $E^-_{test}$ test samples from the 
test split $E_{test}$. Consequently, we use $E^+_{train} \cup E^-_{train}$ for training our model selection and use $E^+_{test} \cup E^-_{test}$ to compute the AUC evaluation metric. We report results averaged over $10$ runs along with their standard deviations in Table~\ref{tab:linkpred}.
Our node embeddings based on spectral-biased walks outperform the state 
of the art methods with significant margins on most of the datasets. 
Our method better captures not only the adjacent nodes with structural similarity, but also the ones that are farther out, due to our walk's tendency to bias such nodes, and hence pack more such nodes in the context window.

Among the baselines, we find that SEAL~\cite{zhang2018link} and WYS~\cite{abu2018watch} have comparable results for few datasets such as SEAL performs better on dense than sparse datasets and an argument can't be generalized for WYS since its performance is better only for few datasets and not to any specific kind of datasets.

\subsection{Sensitivity Analysis}
We test sensitivity towards the following three hyper-parameters. 
Namely, the spectral-biased walk length $W$, the context window size $C$, and the regularization 
parameter $\gamma$ in our Wasserstein regularizer. 
We measure the AUC (y-axis) by varying $W$ and $C$ over two values each, namely $\{50,100\}$ and $\{5,10\}$, respectively, spanning across four different values of $\gamma$ (in x-axis), as shown in Figure~\ref{fig:senstivity}. We conducted the sensitivity analysis on two \emph{dense} datasets (i.e., Celegans and USAir) and on two \emph{sparse} datasets (i.e., Power and Road-minnesota).

Our accuracy metrics lie within a range of $2\%$ and are always better than baselines (WYS and Node2vec), i.e., are robust to various settings of hyper-parameters. Furthermore, even with shorter walks ($W=50$), our method boasts a stable AUC, indicating that our expected hitting times to structurally similar nodes is quite low in practice.  

\subsection{Node Classification}
\begin{table}
	\caption{Node classification results in accuracy (\%). Bold indicate best and underline indicate second best results.}
	\centering
	\small
	\setlength\tabcolsep{3.5pt}
	\begin{tabular}{lllll}
		\toprule
		Algorithms & DeepWalk & Node2Vec  & \textbf{Our Method} \\
		\hline
		
		Citeseer & 41.56 $\pm$ 0.01 & \underline{42.60 $\pm$ 0.01} & \textbf{51.8 $\pm$ 0.25}  \\
		Cora & 66.54 $\pm$ 0.01 & \underline{67.90 $\pm$ 0.52} & \textbf{70.4 $\pm$ 0.30} \\
		Pubmed & 69.98 $\pm$ 0.12 & \underline{70.30 $\pm$ 0.15}  & \textbf{71.4 $\pm$ 0.80} \\

		\bottomrule
	\end{tabular}
	\label{tab:node}
\end{table}
In addition to link prediction, we also demonstrate the efficacy of our node embeddings, via node classification experiments on three citation networks, namely Pubmed, Citeseer, and Cora.
We produce node embeddings from our algorithm and perform classification of nodes without taking node attributes into consideration. We ran experiments on the train-test data splits already provided by~\cite{yang2016revisiting}. 
Results are compared against \emph{Node2vec} and \emph{Deepwalk}, as other state-of-the-art methods for node classification assumed auxiliary node features during training. 
Results in Table \ref{tab:node} show that our method beats the baselines.

\subsection{Effect of Probability, $p$}
\begin{figure}[tbp]
	\centering
	\includegraphics[width=0.8\linewidth]{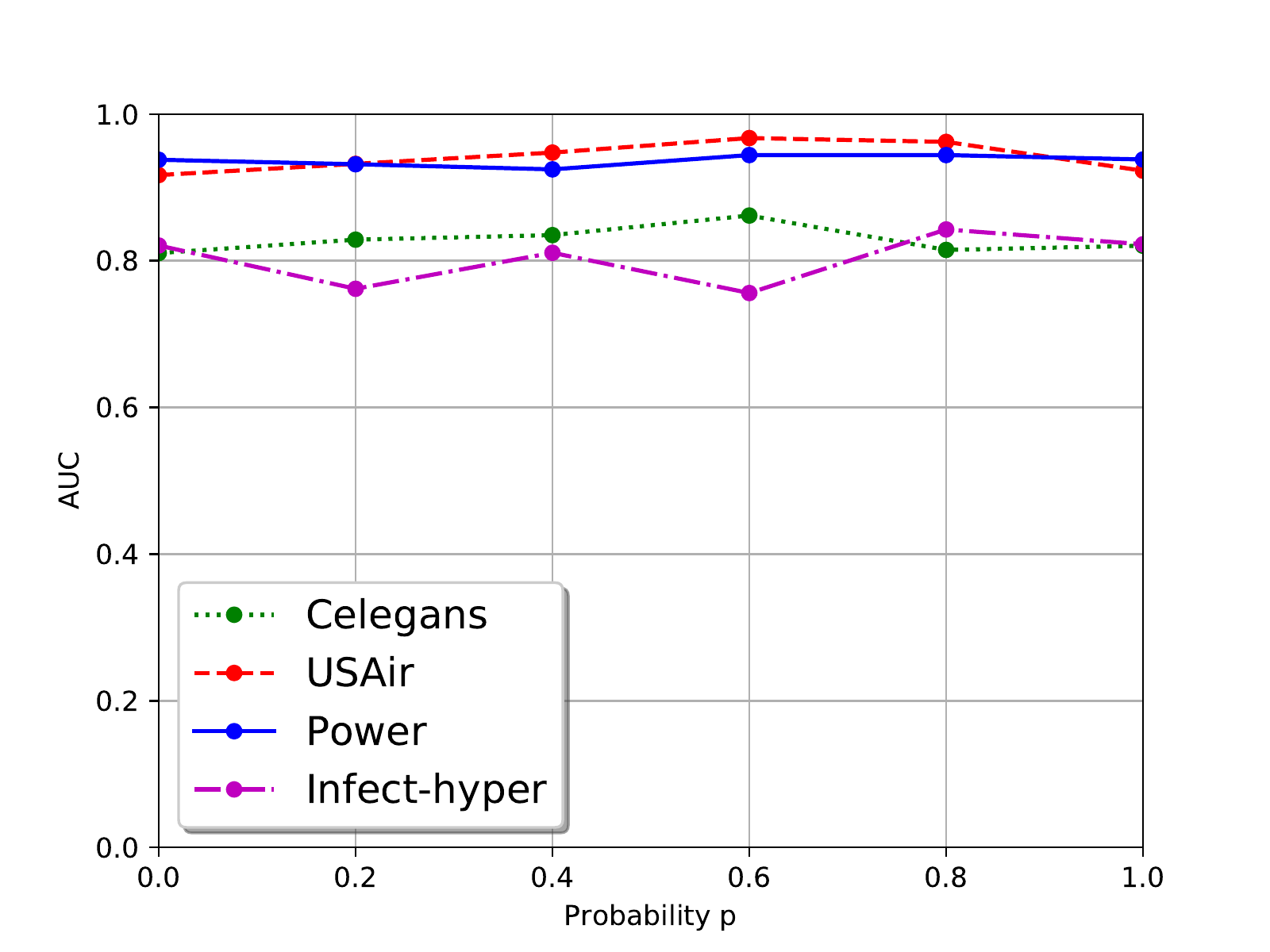}
	\caption{Effect of $p$ on AUC for four datasets for link prediction task.}
	\label{fig:p}
\end{figure}
In earlier sections of the paper, we showed that our algorithm picks the next node in the walk 
from nodes with similar neighborhoods, with probability $p$.
Thus, we conducted an experiment to show the effect of $p$ on the final result (AUC) of 
link prediction. Here, $p$ ranges from $0$ to $1$, 
where $p=0$ implies that the next node is picked completely at random from the 1-hop 
neighborhood (as in simple random walk) and $p=1$ indicates that every node is picked
from the top-$k$ structurally similar nodes in the neighborhood. 

As we move towards greater values of $p$, we tend to select more spectrally similar nodes in the walk. Results are shown in Figure \ref{fig:p} for four datasets Celegans, USAir, Power and Infect-hyper. Figure \ref{fig:p} shows that there is an improvement in AUC for $5\%$ and $2\%$ on an average for dense and sparse datasets respectively. Increase in 
AUC is recorded when $p$ increases up to a certain value of $p$ ranges between $0.4$ to $0.8$.

\section{Conclusions}
We introduced node embeddings based on \emph{spectral-biased random walks}, rooted in an \emph{awareness} of the neighborhood structures surrounding the visited nodes. 
We further empirically studied the quality of the spectral-biased random walks by comparing their 
expected hitting time between pairs of spectrally similar nodes, packing density of fixed-sized walks, 
and the cover time to hit all the spectrally similar nodes within a fixed Wasserstein ball defined by us.
We found our spectral-biased walks outperformed simple random walks in all the aforementioned quality parameters.

Motivated by our findings and in an attempt to break away from word vector models, we proposed a paragraph vector model along with a novel Wasserstein regularizer. Experimentally, we showed that our method significantly outperformed existing state-of-the-art node embedding methods on a large and diverse set of graphs, for both link prediction and node classification.

We believe that there does not exist a ``one-size-fits-all" graph embedding for all applications and domains. Therefore, our future work will primarily focus on \emph{generalizing} our biased walks to a broader class of functions that could possibly capture graph properties of interest to the applications at hand.

\bibliographystyle{./IEEEtran}
\bibliography{./IEEEabrv,./linkpred}
\clearpage
\appendix
\section{Appendix}
\subsection{Statistics of spectral-biased walk}
In order to understand the importance of our spectral-biased random walk, we perform few more experiments similar to those mentioned in our main paper. We conducted mainly two experiments based on walk length $W$ and probability $p$ for spectral-biased 
walk (SW) and random walk (RW).
\paragraph{Spectrally similar nodes}
We compare our walk, SW with RW in terms of covering spectrally similar nodes in walks of varying lengths from $40$ to $200$. We randomly sample $100$ nodes in each of the six datasets, generate both the walks with walk length, $W = \{40, 80, 120, 160, 200\}$ 
and average it over $100$ runs. Both the walks are compared with the percentage of packing spectrally similar nodes in their walks. We computed results of SW for two values of probability $p$. Figure~\ref{fig:spec_sim} shows the plots for six 
datasets considering both sparse and dense datasets. Our plots show that our walk, SW packs more number of spectrally similar nodes in its walk even for different values of $p$ than random walk, RW. In addition, our walk with shorter length is still able to outperform RW. 
We also find that some datasets have large margins between both the walks in covering spectrally similar nodes like Celegans, Power, etc. However, SW always covers more spectrally similar nodes in the walk than RW for all the datasets.

\begin{algorithm}
	\caption{decides which vertex to transition to next
		\label{alg:move_to}}
	\begin{algorithmic}[1]
		\Require{vertex $v$, $A \in \R^{n \times n}$}
		\Statex
		\Function{move\_to}{$v, A$}
		
		
		\State - Compute transition probability vector 
		\hspace{1em}\Let{$[p(v_1), \dots ,p(v_n)] $ } {$A[v]$}
		
		\State - Divide $[0,1]$ into intervals 
		\State \hspace{.5em}$I_j := F(v_{j -1}),F(v_j)]$ for $j=(1,\dots n)$
		\State \hspace{.5em}where $F(v_i) = \sum_{j=1}^{i} p(v_j)$, $F(v_0)=0$
		
		\State - Find $j$ s.t. random number $x \sim [0,1] \in I_j$
		\State \Return $v_j$
		\EndFunction
	\end{algorithmic}
\end{algorithm}
\subsection{Algorithms}
In this section, we describe in detail the algorithms used. 
Algorithm~\ref{alg:move_to} serves as a helper sub-routine to Algorithm~\ref{alg:spec_rw} 
to decide the next vertex to move to depending on the transition matrix passed to it as 
input.

Our algorithm is initiated with the given transition matrices, walk length, parameter $\epsilon$ and an initial vertex $v_0$. In each iteration, the current vertex $v_{curr}$ take a move to the next adjacent vertex $v_{next}$ using one of 
the transition matrices $W$ and $P$. This 
decision is based on the value of $x$ taken uniformly at random and compared with the probability parameter $\epsilon$. With probability $\epsilon$, the next adjacent vertex would be picked up using biased transition matrix $W$, and 
with probability $1-\epsilon$, next adjacent vertex is taken uniformly at random using transition matrix $P$.
\begin{algorithm}
	\caption{Spectral-Biased Random Walk
		\label{alg:spec_rw}}
	\begin{algorithmic}[1]
		\Require{initial vertex $v_0$, transition matrices $P,W \in \R^{n \times n}$, walk length $k$, parameter $\epsilon$}
		\Statex
		\Function{spectral\_walk}{$v_0, P,W,k,\epsilon$}
		
		\Let{$v_{curr}$}{$v_0$}
		\For{$i \gets 1 \textrm{ to } k-1$}
		\State $x \sim \mU[0,1]$
		\If{$x  \leq \epsilon$}
		\Let{$ v_{next} $ } {move\_to($v_{curr}, W$) }
		\Else
		\Let{$ v_{next} $ } {move\_to($v_{curr}, P$) }					
		\EndIf			
		\Let{ $v_{curr}$ } { $v_{next}$ }
		\Let{ Walk $w$ } { append $v_{curr}$ to $w$}
		\EndFor
		\State \Return $w$
		\EndFunction
	\end{algorithmic}
\end{algorithm}

\subsection{Training}

The training setup of our method is explained in the main paper where parameters are as follows: probability $p=0.6$, walk length $W=100$ with $50$ walks per node, context window size $C=10$, 
regularization term $\gamma$ ranges from $1e-6$ to $1e-8$, node embedding dimension is $128$ with $100$ to $200$ epochs to train the model. We observe that our model is quite fast, including the time for pre-processing step of spectral-biased walk generation. 
 We also analyze sensitivity of hyper-parameters in Figure 2 in main paper and Figure~\ref{fig:senstivity1} to show the robustness 
of our algorithm. Along with 
sensitivity, we also discuss how probability $p$ affects the quality of our walk and finally show the results of our method in Figure 3 in main paper. In order to understand the importance of our spectral-biased walk, we perform few experiments for which results are 
shown in Figures~\ref{fig:spec_sim} and \ref{fig:early_y}.

\subsection{Ablative Study}\label{sec:ablative}
In this section, we illustrate the results of link prediction task for $50\%-50\%$ train-test split in Table \ref{tab:5050}. Results for $90\%-10\%$ split is provided in Table I of main paper. The observation we can draw here is that with partial data of randomly picking 
$50\%$ existing (positive) links and $50\%$ non-existent (negative) links for training, we outperform existing methods for majority of the datasets. We can see that dense datasets are not affected by large margins whereas sparse datasets have an effect of partial data in 
the results. From the baseline methods, SEAL is comparable to our results for almost $6/10$ datasets. 

While WYS shows more stable AUC results (with no drops) for sparse datasets, it still suffers from a huge standard deviation in its reported AUC values and is hence not as stable. We also observe that WLNM and WLK perform comparably to SEAL and our method, for sparse datasets. But, the drawback of kernel based methods is that they are not efficient in 
terms of memory and computation time, as they compute pairwise kernel matrix between the subgraphs of all the nodes in the graph.
\begin{table*}
	\caption{Link prediction results with 50\% training links. "-" represents incomplete execution due to either \emph{out of memory errors} or computation time exceeding $20$ hours. Bold indicate best and underline indicate second best results.}
	\centering
	\footnotesize
	\setlength\tabcolsep{3.5pt}
	\begin{tabular}{llllllll}
		\toprule
		Algorithms & Node2Vec & VGAE & WLK & WLNM & SEAL & WYS & \textbf{Our Method} \\
		\hline
		
		Power & 52.24 $\pm$ 0.31 & 51.26 $\pm$ 0.87 & - & - & \underline{59.91 $\pm$ 0.12} &  \textbf{88.14 $\pm$ 10.43} &  53.22 $\pm$ 0.05  \\
		Celegans & 58.98 $\pm$ 0.43 & 76.50 $\pm$ 0.79 & 66.58 $\pm$ 0.29 & 65.99 $\pm$ 0.34 & \underline{82.35 $\pm$ 0.41} &  74.67 $\pm$ 8.01 & \textbf{83.22 $\pm$ 0.11}  \\
		USAir & 75.21 $\pm$ 0.59 & 92.00 $\pm$ 0.10 & 83.49 $\pm$ 0.39 & 84.39 $\pm$ 0.48 & \underline{95.14 $\pm$ 0.10} & 93.81 $\pm$ 3.65 & \textbf{95.31 $\pm$ 0.35} \\
		Road-Euro & 51.27 $\pm$ 0.98 & 48.53 $\pm$ 0.40 & 60.37 $\pm$ 0.41 & \underline{63.61$\pm$ 0.52}& 49.29 $\pm$ 0.92 & \textbf{77.56 $\pm$ 20.33} & 52.09 $\pm$ 0.15\\ 
		Road-Minnesota & 50.94 $\pm$ 0.67 &50.72 $\pm$ 1.11 & \underline{68.15 $\pm$ 0.44} & 67.18 $\pm$ 0.41 & 57.43 $\pm$ 0.13 &  \textbf{76.07 $\pm$ 20.12} & 53.65 $\pm$ 0.21 \\
		Bio-SC-GT & 82.21 $\pm$ 0.54 & 85.64 $\pm$ 0.21 & - & - & \underline{96.07 $\pm$ 0.03} & 87.84 $\pm$ 7.01 & \textbf{96.54 $\pm$ 0.12}\\
		Infect-hyper & 61.38 $\pm$ 0.28 & 76.29 $\pm$ 0.20 & 60.99 $\pm$ 0.59 & 63.60 $\pm$ 0.61 & 75.11 $\pm$ 0.05 & \underline{78.32 $\pm$ 0.31} & \textbf{81.34 $\pm$ 0.24} \\
		PPI & 60.14 $\pm$ 0.18 & 88.60 $\pm$ 0.11 & - & - & \textbf{91.47 $\pm$ 0.04} & 85.01 $\pm$ 0.32 & \underline{90.38 $\pm$ 0.18} \\
		Facebook & 95.93 $\pm$ 0.11 & - & - & - & 80.11 $\pm$ 0.23 & \underline{98.78 $\pm$ 0.10} & \textbf{99.13 $\pm$ 0.01} \\
		HepTh & 75.21 $\pm$ 0.21 & 82.60 $\pm$ 0.25 & - & - & \underline{90.47 $\pm$ 0.07} & \textbf{93.45 $\pm$ 3.67} & 90.13 $\pm$ 0.03 \\
		\bottomrule
	\end{tabular}
	\label{tab:5050}
\end{table*}

\subsection{Sensitivity Analysis}
\begin{figure*}
  \centering
  \begin{subfigure}[b]{0.4\textwidth}
    \includegraphics[width=\linewidth]{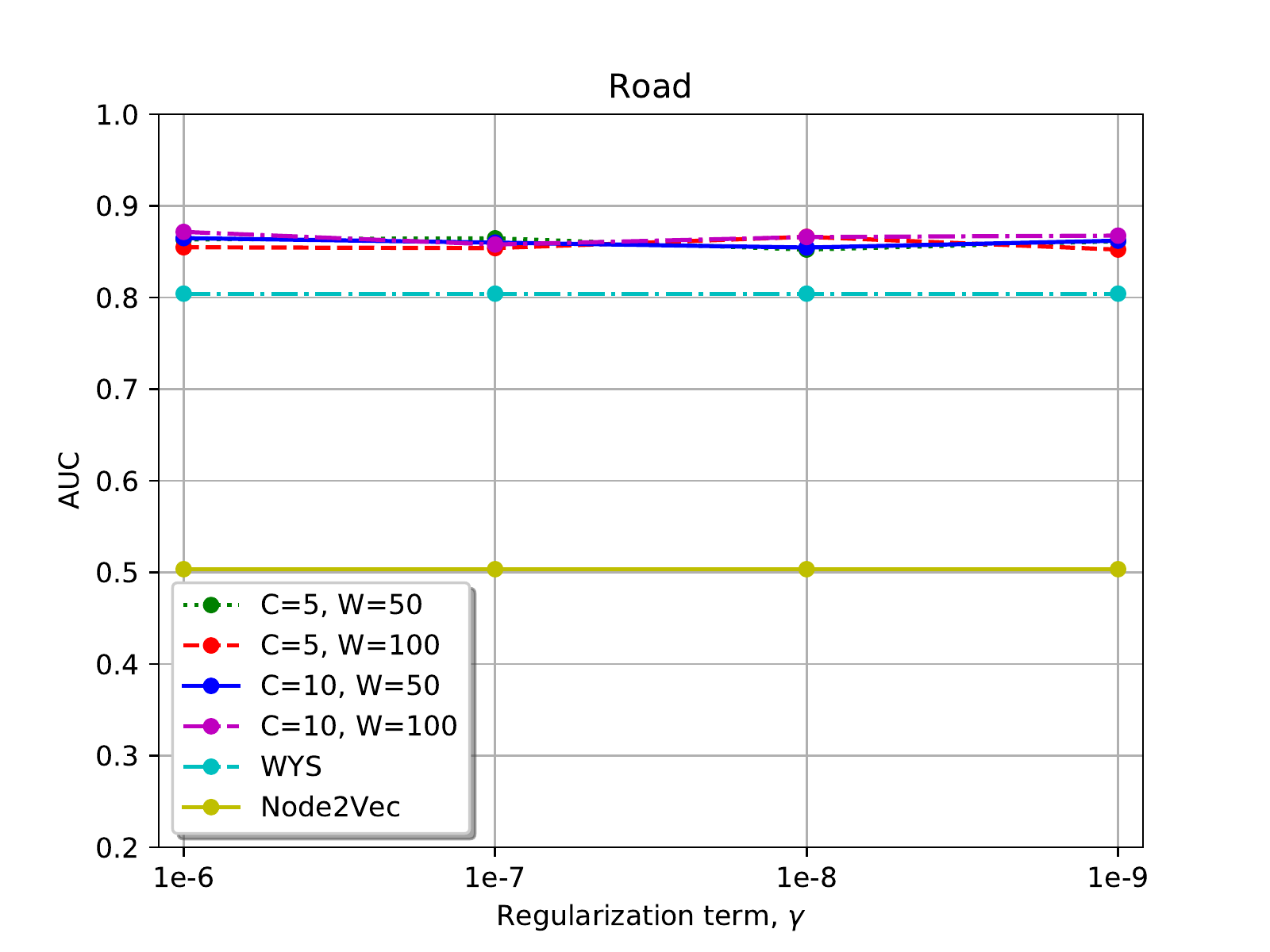}
  \end{subfigure}%
  \begin{subfigure}[b]{0.4\textwidth}
    \includegraphics[width=\linewidth]{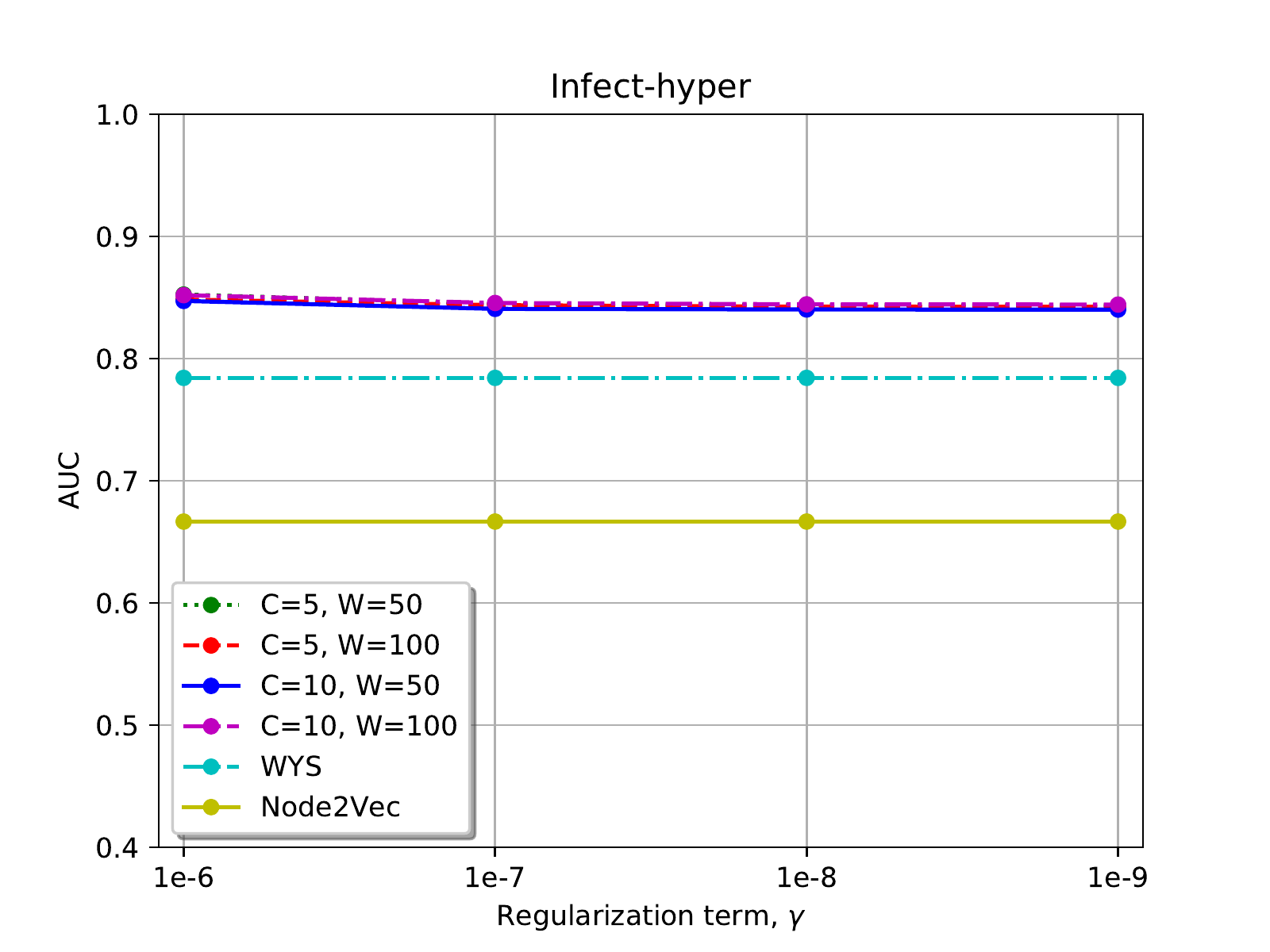}
  \end{subfigure}%
       \\
  \begin{subfigure}[b]{0.4\textwidth}
    \includegraphics[width=\linewidth]{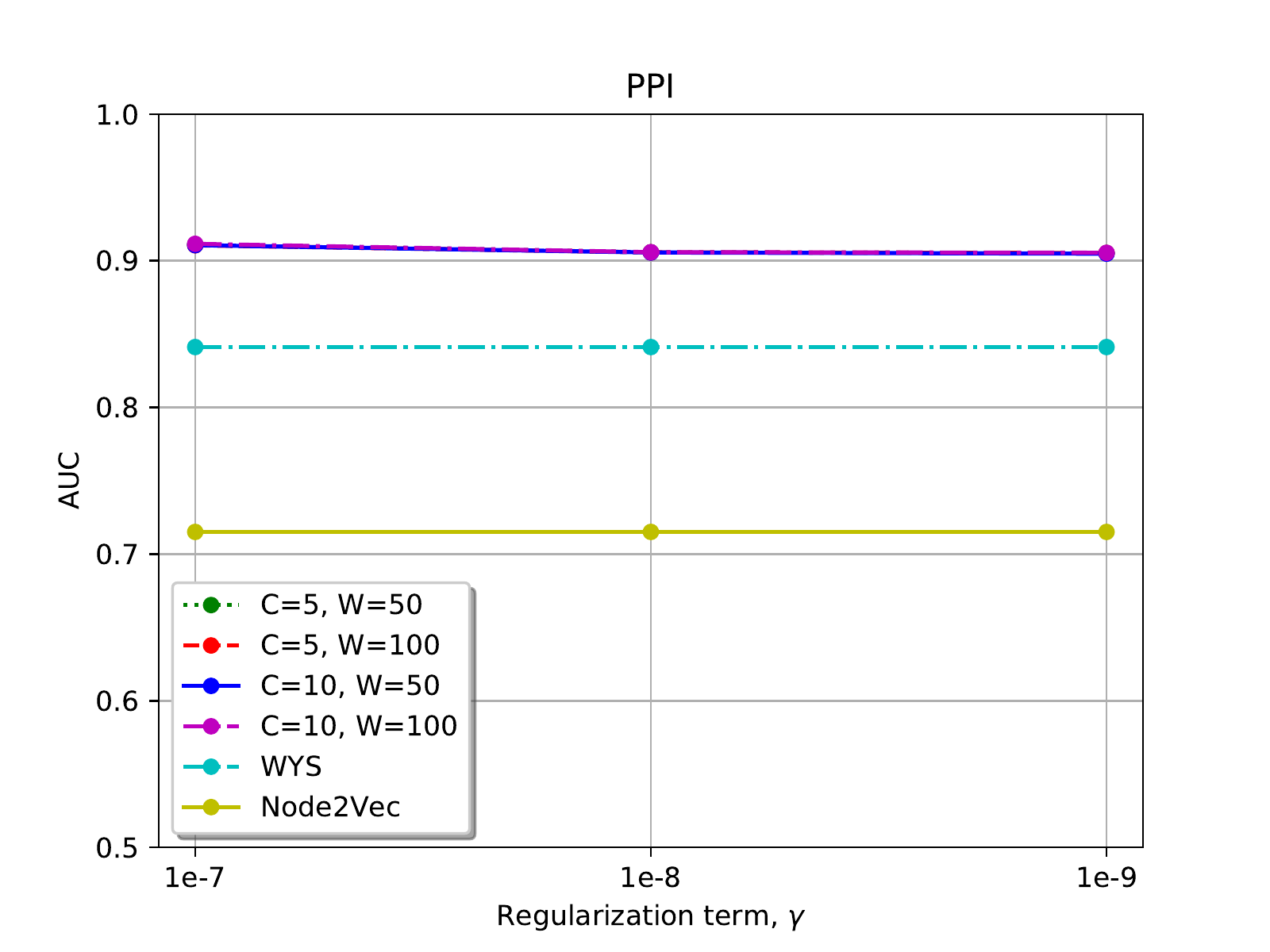}
  \end{subfigure}%
  \begin{subfigure}[b]{0.4\textwidth}
    \includegraphics[width=\linewidth]{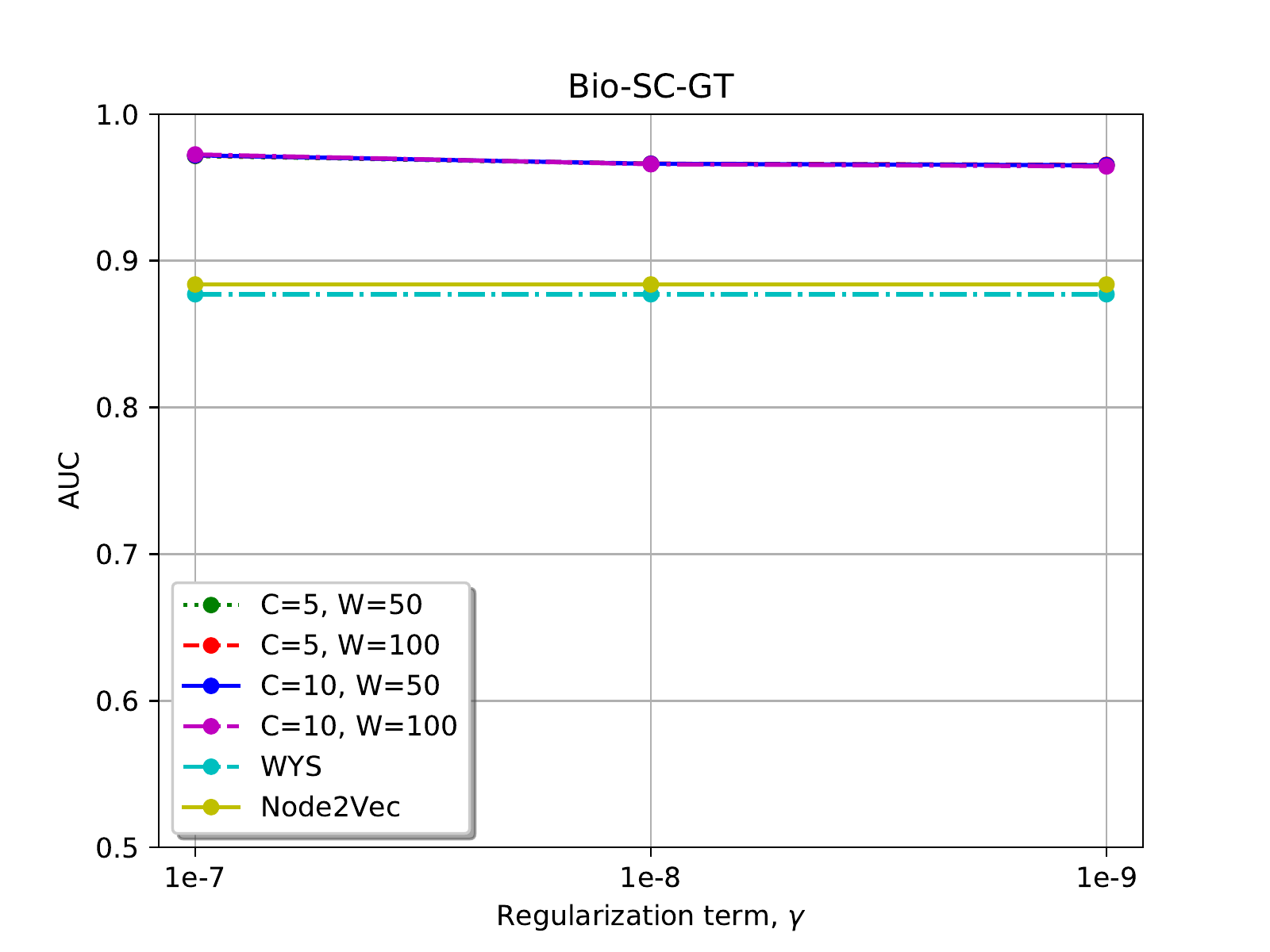}
  \end{subfigure}%
  \caption{Sensitivity of window size, $C$ and walk length, $W$ with respect to regularization term, $\gamma$ is measured in AUC for four datasets for link prediction task.}
  \label{fig:senstivity1}
\end{figure*}

\begin{figure*}
  \centering
  \begin{subfigure}[b]{0.5\textwidth}
    \includegraphics[width=\linewidth]{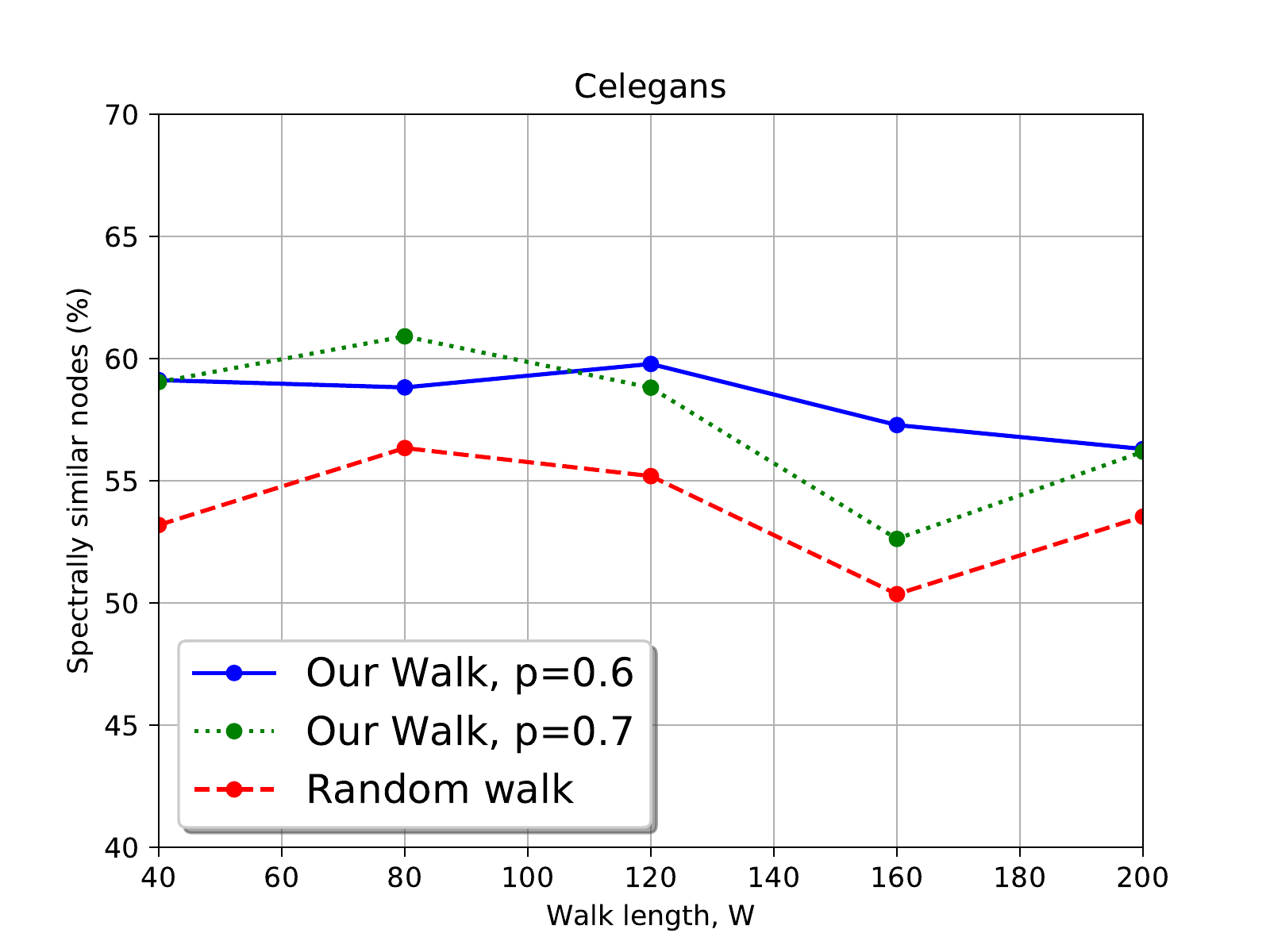}
  \end{subfigure}%
  \begin{subfigure}[b]{0.5\textwidth}
    \includegraphics[width=\linewidth]{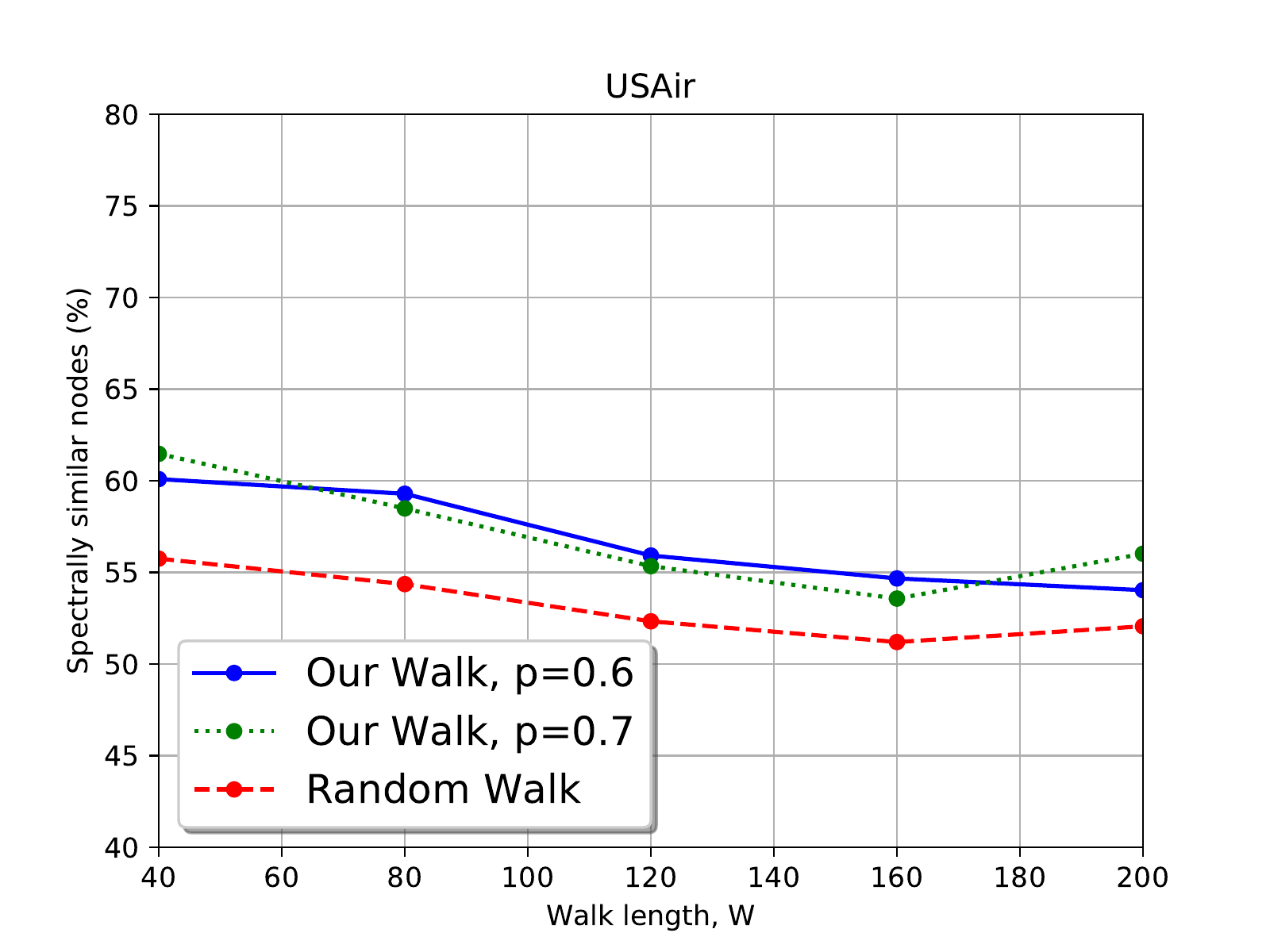}
  \end{subfigure}%
       \\
  \begin{subfigure}[b]{0.5\textwidth}
    \includegraphics[width=\linewidth]{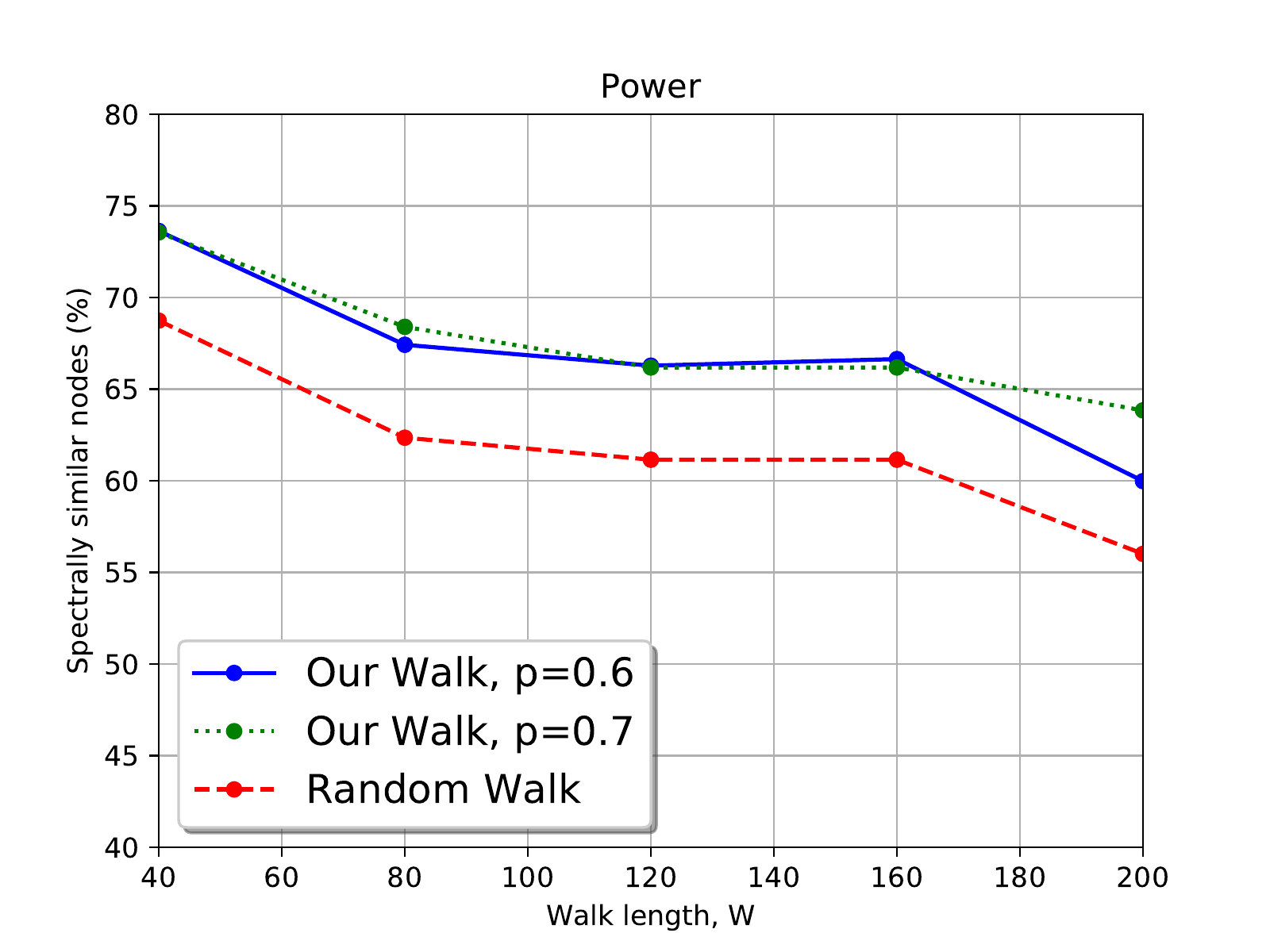}
  \end{subfigure}%
  \begin{subfigure}[b]{0.5\textwidth}
    \includegraphics[width=\linewidth]{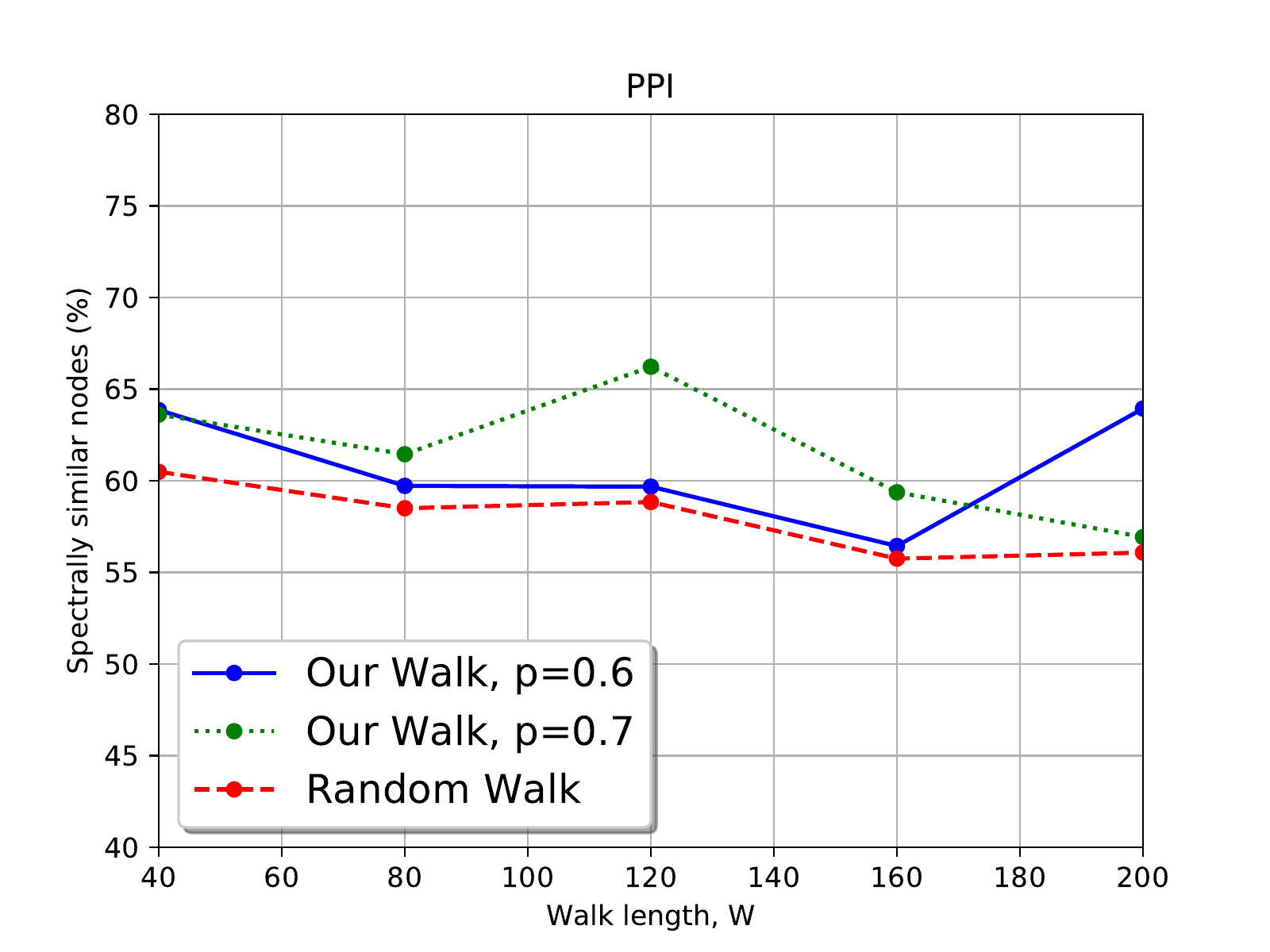}
  \end{subfigure}%
  \\
    \begin{subfigure}[b]{0.5\textwidth}
    \includegraphics[width=\linewidth]{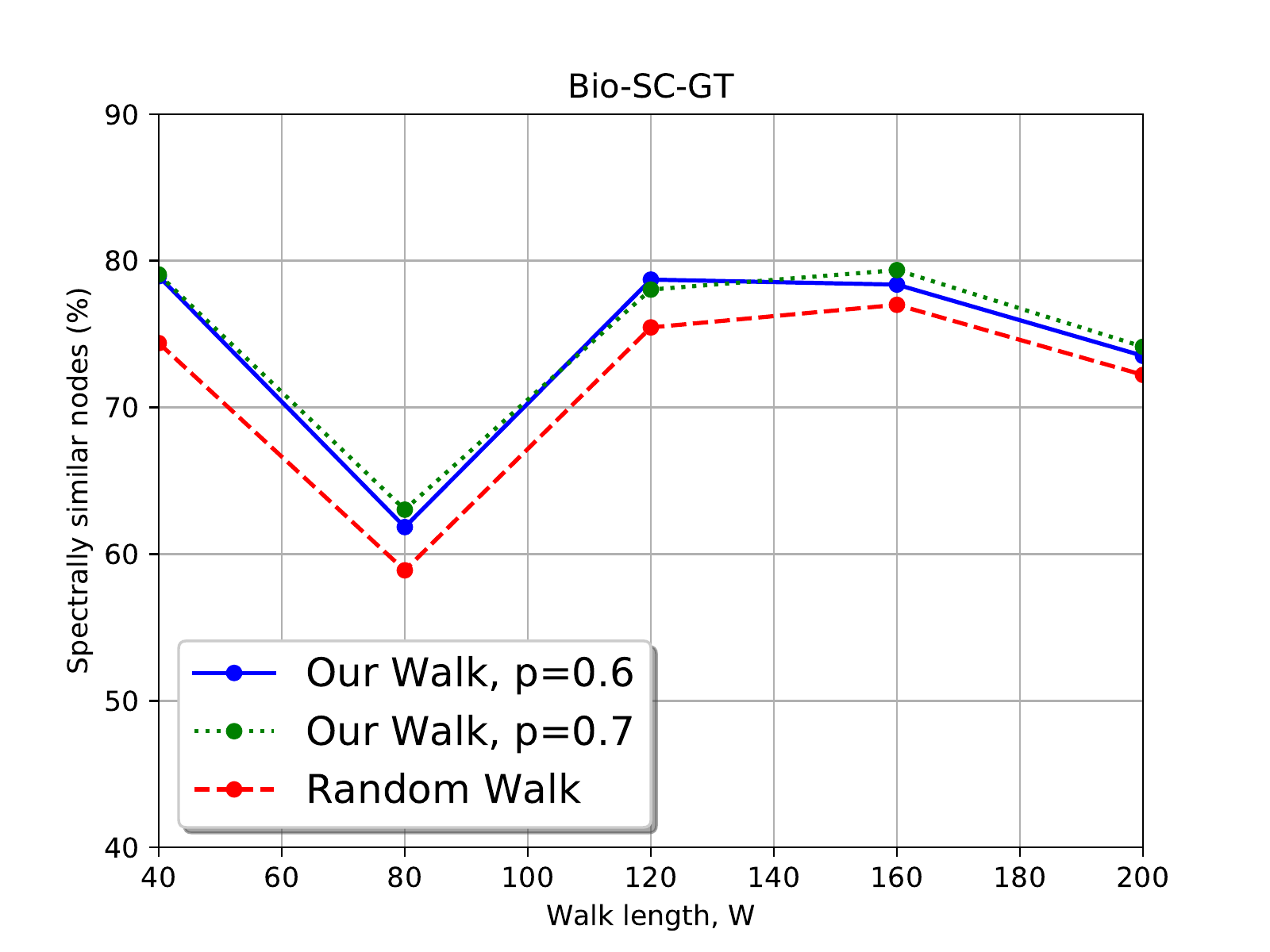}
  \end{subfigure}%
    \begin{subfigure}[b]{0.5\textwidth}
    \includegraphics[width=\linewidth]{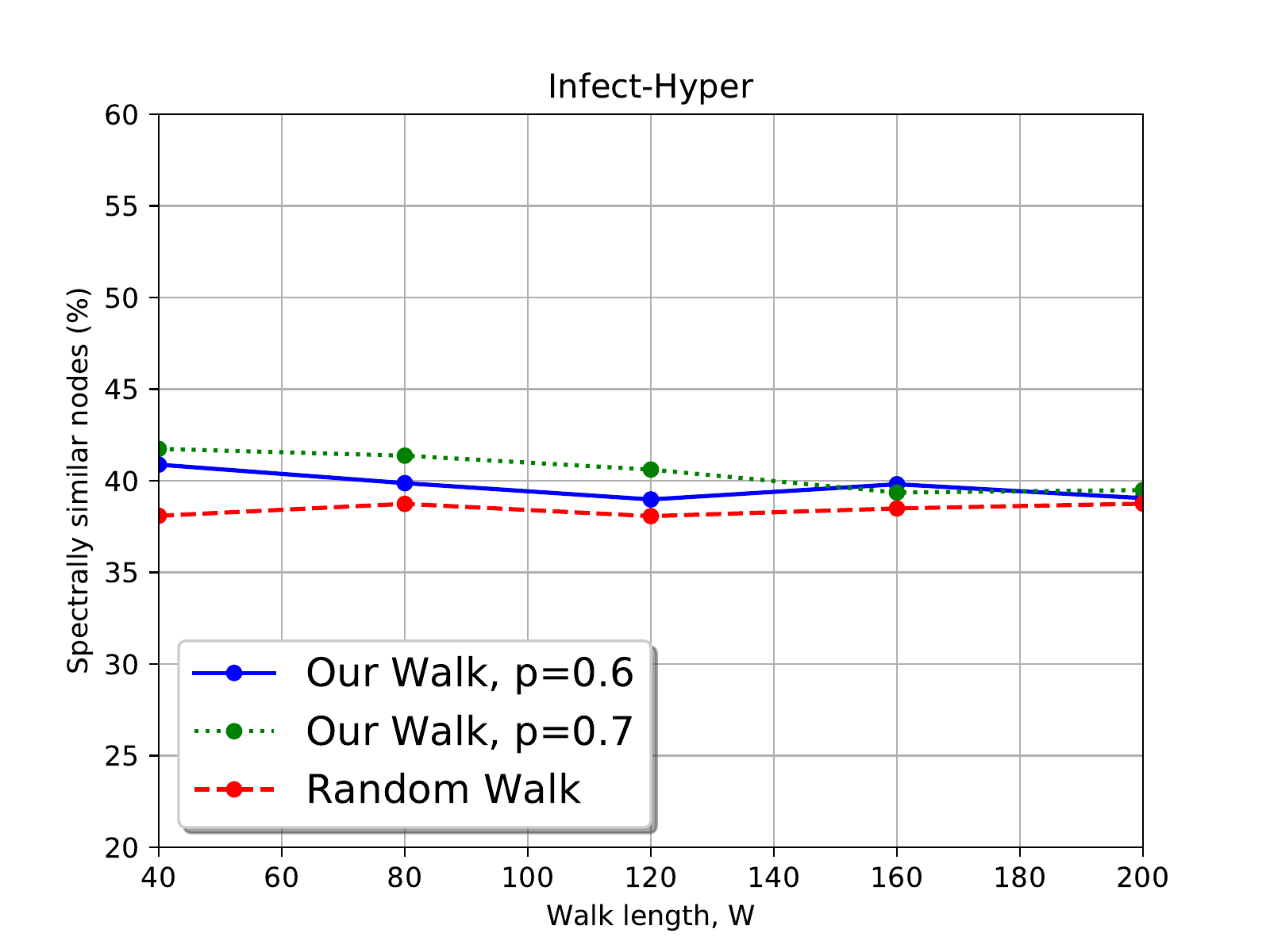}
  \end{subfigure}%
  \caption{Percentage of spectrally-similar nodes packed in walks of varying length for random walk (RW) and spectral-biased walk (SW) for six datasets.}
  \label{fig:spec_sim}
\end{figure*}

For sensitivity, we run our algorithm for three different hyper-parameters in order to check robustness of our method. These hyper-parameters are spectral-biased walk length $W$, the context window size $C$ and the regularization term $\gamma$ for our 
Wasserstein regularizer. We perform experiments on eight datasets for link prediction, out of which results for four datasets are mentioned in our main paper. Figure~\ref{fig:senstivity1} shows the results for four datasets with $C=5,10$, $W=50,100$ and 
regularization term on $x$-axis. Here, Road is a sparse dataset and other three datasets (Infect-Hyper, Bio-SC-GT and PPI) are dense datasets. We see that our results lie within the range of $2\%$ and are stable for different hyper-parameters. We observe that our results are always above the baseline methods (WYS and Node2Vec).

\begin{figure*}
  \centering
  \begin{subfigure}[b]{0.5\textwidth}
    \includegraphics[width=\linewidth]{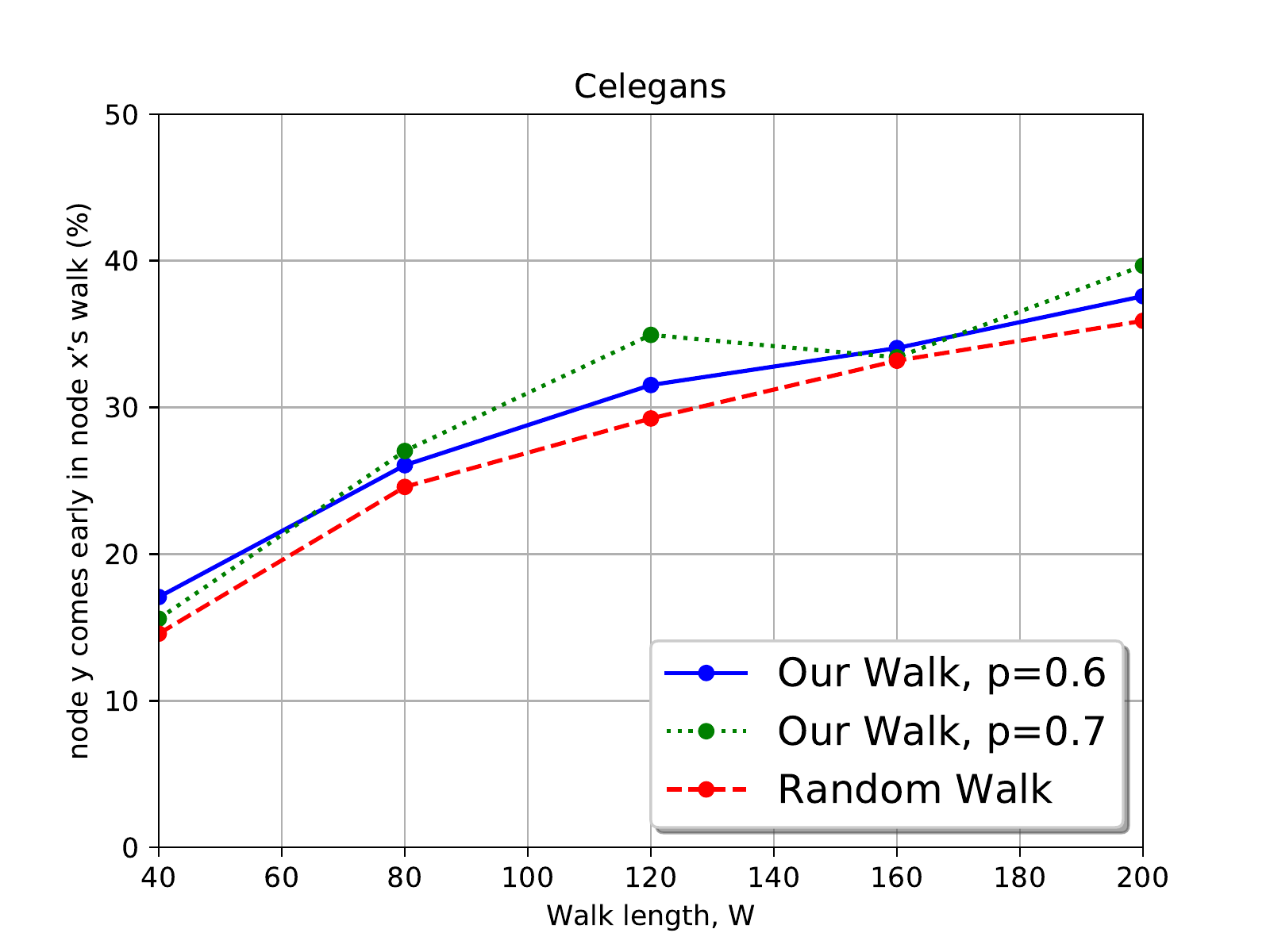}
  \end{subfigure}%
  \begin{subfigure}[b]{0.5\textwidth}
    \includegraphics[width=\linewidth]{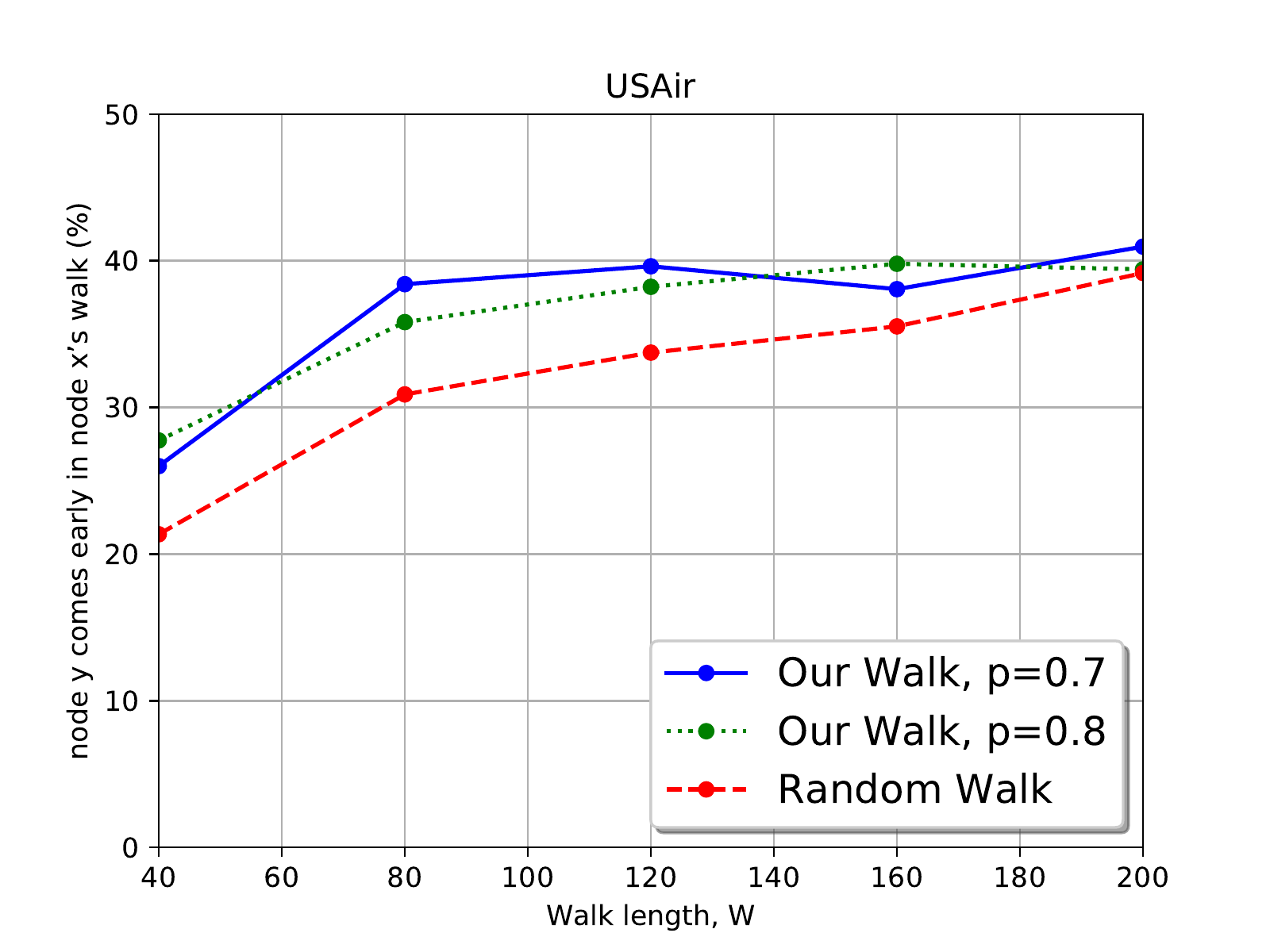}
  \end{subfigure}%
       \\
  \begin{subfigure}[b]{0.5\textwidth}
    \includegraphics[width=\linewidth]{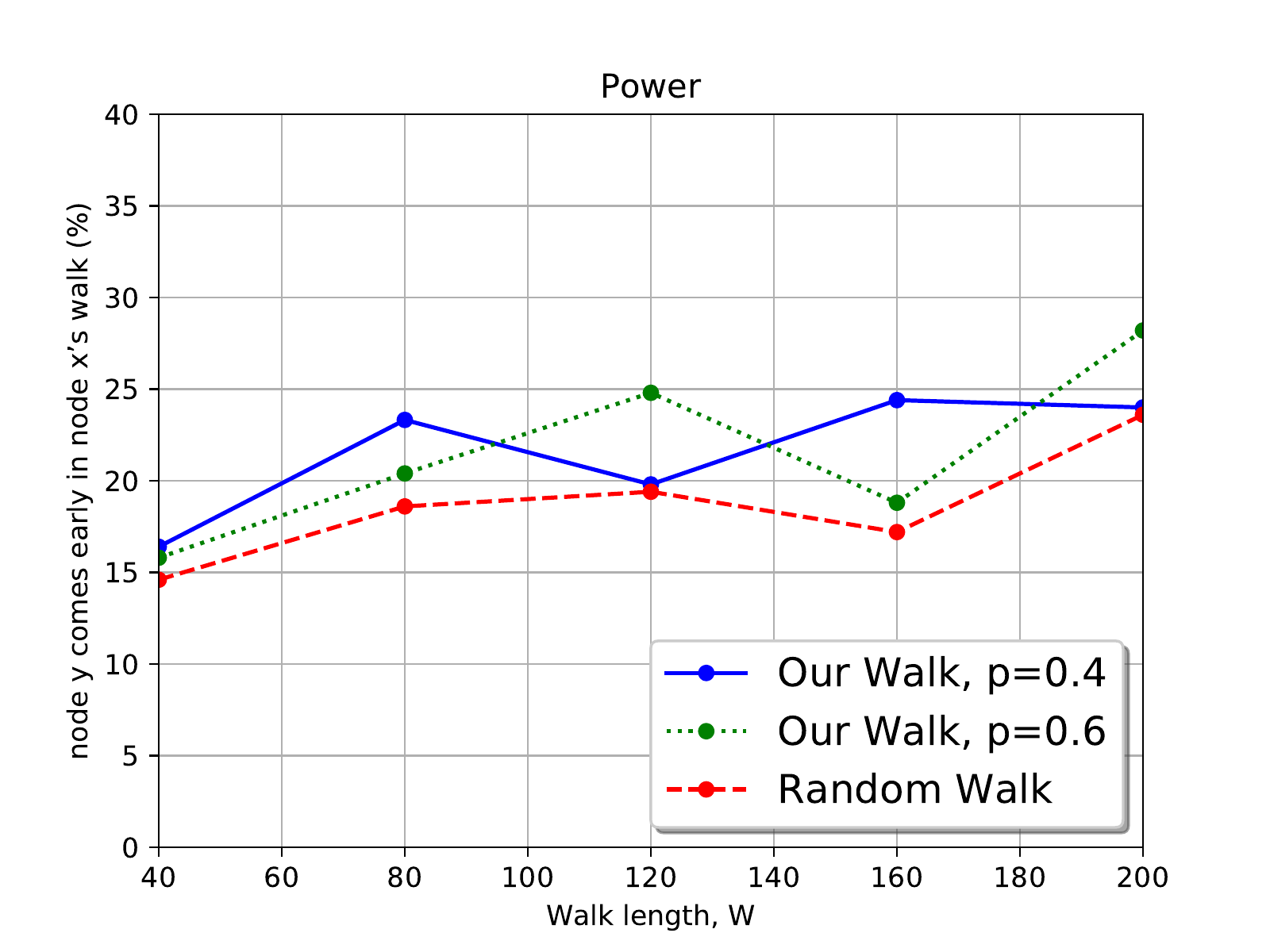}
  \end{subfigure}%
  \begin{subfigure}[b]{0.5\textwidth}
    \includegraphics[width=\linewidth]{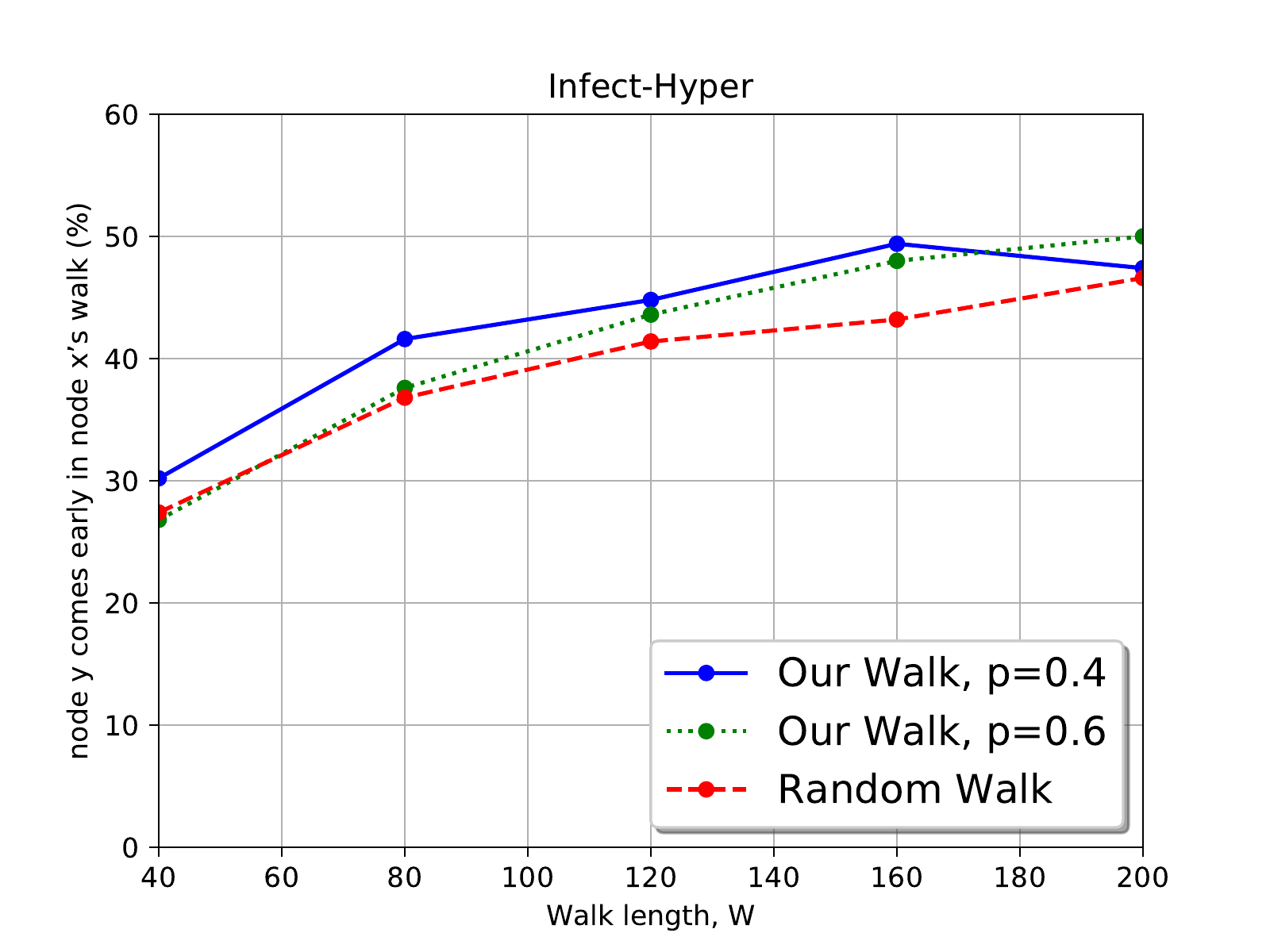}
  \end{subfigure}%
  \\
    \begin{subfigure}[b]{0.5\textwidth}
    \includegraphics[width=\linewidth]{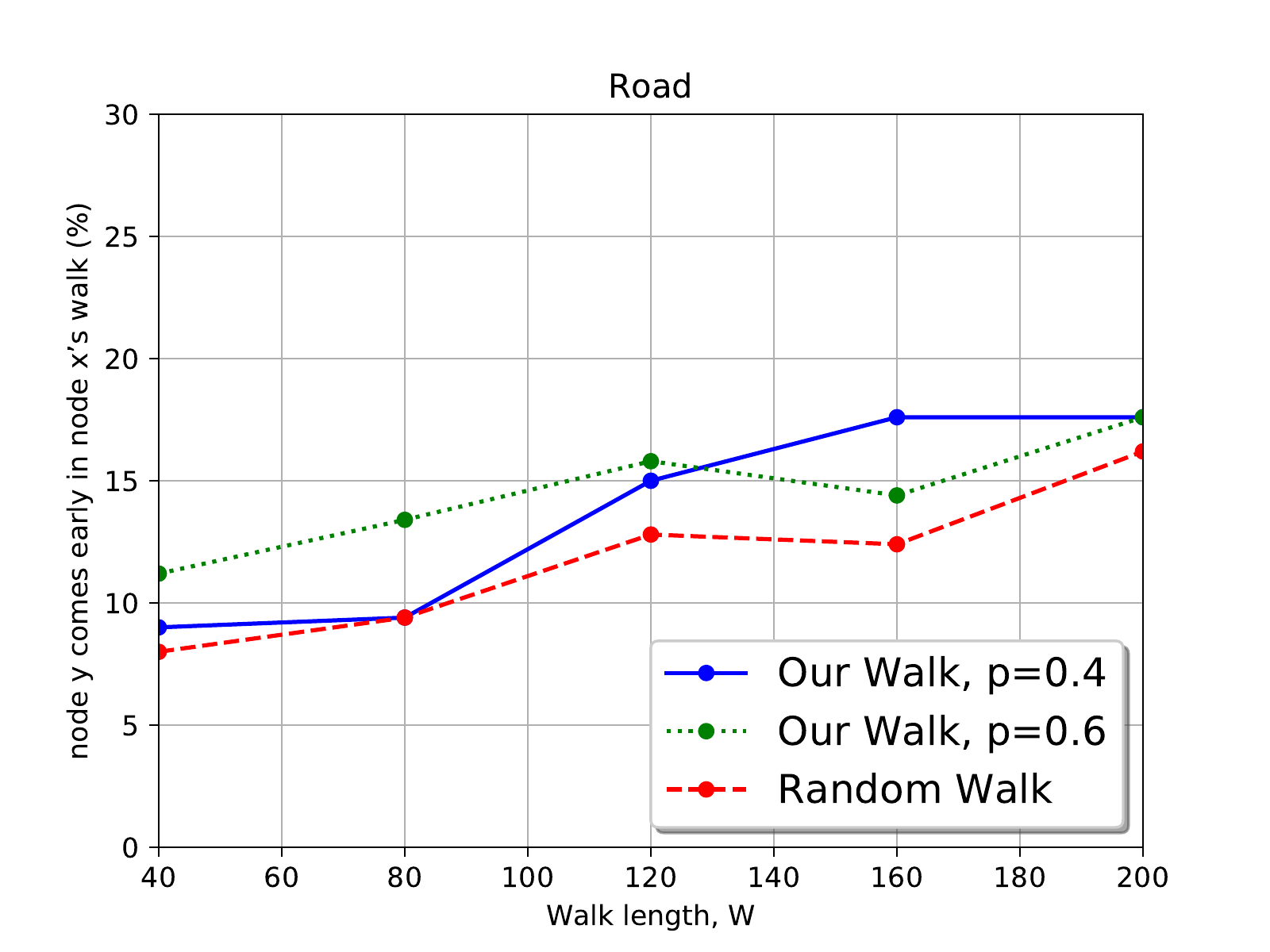}
  \end{subfigure}%
    \begin{subfigure}[b]{0.5\textwidth}
    \includegraphics[width=\linewidth]{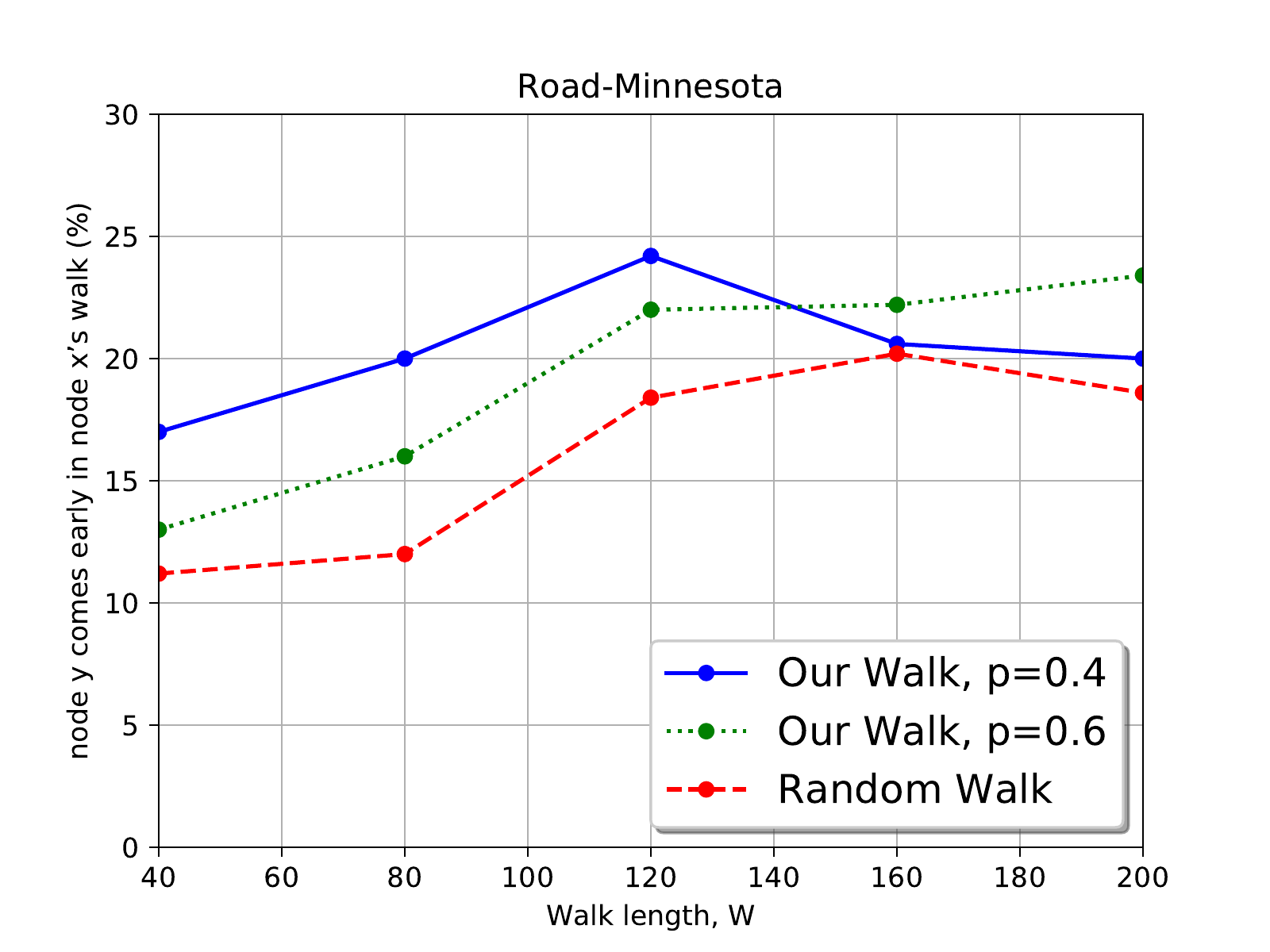}
  \end{subfigure}%
  \caption{Average ranking of target nodes encountered by simple random walk (RW) and our spectral-biased walk (SW) of varying length for six datasets.}
  \label{fig:early_y}
\end{figure*}
\paragraph{Ranking of target nodes}
We also studied the hitting time of target node $j$ to appear in the walk when a walk is initiated from a node $i$ for both the walks SW and RW. For this experiment, we pick $100$ unseen pair of nodes $(i,j)$. Both the walks are generated starting 
from node $i$ for varying length from $40$ to $200$, considering two values of probability $p$ for our walk. We compute the number of times $j$ appears early in $i$'s walk for SW and RW and we average the results over $100$ runs. We conducted this experiment for six datasets covering both sparse and dense datasets.
The results are shown in Figure~\ref{fig:early_y}. The analysis from our results represents the lower hitting time for our walk (i.e., SW) in ranking of a target node appearance in a walk starting from a node $i$ on all the datasets. We found that the 
percentage of target nodes appears early in SW is more than in RW.

\end{document}